\RequirePackage{fix-cm}
\documentclass[twocolumn]{svjour3}          % twocolumn
\smartqed  % flush right qed marks, e.g. at end of proof
\usepackage{graphicx,comment}
\usepackage{amsmath,algorithm, url,amsfonts,graphicx,color,bbm,ulem,dsfont,subcaption}
\begin{document}

\title{Spatially Constrained Spectral Clustering Algorithms for Region Delineation %\thanks{Grants or other notes
}
%\subtitle{Exploring the Trade-off between Spatial Contiguity and Landscape Homogeneity}


\author{Shuai Yuan$^1$ \and
        Pang-Ning Tan$^1$ \and
        Kendra Spence Cheruvelil$^2$  \and
        Sarah M. Collins$^3$ \and
        Patricia A. Soranno$^3$        }


\institute{%Shuai Yuan \at
            %  \email{yuanshu2@msu.edu}           %  \\
 %          \and
           1.Dept of Comp Science and Engr, Michigan State University\\
           2.Dept of Fisheries and Wildlife \& Lyman Briggs College, Michigan State University\\
           3.Dept of Fisheries and Wildlife, Michigan State University
}

\date{Received: date / Accepted: date}

\maketitle

\begin{abstract}

Regionalization is the task of dividing up a landscape into homogeneous patches with similar properties. Although this task has a wide range of applications, it has two notable challenges. First, it is assumed that the resulting regions are both homogeneous and spatially contiguous. Second, it is well-recognized that landscapes are hierarchical such that fine-scale regions are nested wholly within broader-scale regions. To address these two challenges, first, we develop a spatially constrained spectral clustering framework for region delineation that incorporates the tradeoff between region homogeneity and spatial contiguity. The framework uses a flexible, truncated exponential kernel to represent the spatial contiguity constraints, which is integrated with the landscape feature similarity matrix for region delineation. To address the second challenge, we extend the framework to create fine-scale regions that are nested within broader-scaled regions using a greedy, recursive bisection approach. We present a case study of a terrestrial ecology data set in the United States that compares the proposed framework with several baseline methods for regionalization. Experimental results suggest that the proposed framework for regionalization outperforms the baseline methods, especially in terms of balancing region contiguity and homogeneity, as well as creating regions of more similar size, which is often a desired trait of regions.

%\PAS{i edited the abstract quite a bit, but i just tried to simplify and clarify a bit to link the two 'challenges' a little better. i don't think i added length. i tried not to}
%\KSC{would it be appropriate to add a sentence here that highlights the result (that your algo is better?)} \SC{yes, or maybe instead of adding you could rephrase to say we compared our algo against others and demonstrate that it is most effective in terms of producing regions etc.} \PN{Results will be added after it is re-written.}

\keywords{ Constrained spectral clustering \and Regionalization \and hierarchical clustering}
\end{abstract}

\section{Introduction} \label{intro}
\label{sec:intro}
A regionalization framework delineates the geographical landscape into spatially contiguous, homogeneous units known as regions or zones. Regionalizations are important because they provide the spatial framework used in many disciplines, including landscape ecology, environmental science, and economics, as well as for applications such as public policy and natural resources management~\cite{cheruvelil2013,Long2010,George1997,Margules1985}. For example, the hierarchical system of hydrologic units described in \cite{Huc1987} provides a standardized regionalization framework that has been widely used in water resource and land use studies~\cite{Duque2007}. Abell et al. \cite{AbellRobin2008} have also developed a global biogeographic regionalization framework that serves as a useful tool for studying biodiversity in freshwater systems and for conservation planning efforts.

McMahon et al.~\cite{mcmahon2001} divide existing multivariate regionalization methods into two categories, qualitative and quantitative. For qualitative methods, regions with similar landscape characteristics are delineated by experts from multiple maps of different geographic features using manual visual interpretation~\cite{bailey1998,omernik1995}. For quantitative methods, clustering approaches such as k-means and hierarchical clustering~\cite{Host1996,Hargrove2004,Jain:1999} are used to partition the geographical area into smaller regions. Although quantitative clustering approaches provide a more systematic and reproducible way to identify regions compared to qualitative approaches, one potential limitation of existing clustering methods is that the regions created may not be spatially contiguous. Region contiguity is a desirable criterion for many applications that treat regions as individual entities for purposes including research, policy, and management  (e.g., site-specific management in precision agriculture~\cite{yan2007}). Therefore, alternative methods are needed that can effectively cluster similar areas based on multiple mapped variables, but have the added constraint of being spatially contiguous.

%The regionalization system should also consider the inherent hierarchical nature of landscapes~\cite{WuJ2013}, in which fine-scale regions are nested wholly within broader-scale regions. This has led to the wide use of constrained-based hierarchical clustering techniques \cite{murtagh1985} such as single-link \cite{numericaleco}, complete-link \cite{rao2003}, UPGMA \cite{Kreft2010,Sandin1999}, and Ward's method \cite{Iyigun2013,Wolock2004} to address the hierarchical region delineation problem. However, as will be shown in this study, such methods tend to generate hierarchies that are highly imbalanced in terms of their region size, and thus, are not that suitable for resource planning and management applications. 

In the preliminary version of this work \cite{Yuan2015}, we presented a spatially constrained spectral clustering framework that uses a truncated exponential kernel~\cite{Kondor2002} to produce spatially contiguous and  homogeneous regions \cite{Kamvar03spectrallearning,BoleyK13,ShiFY2010,Wang2010}. %The framework uses a truncated exponential kernel~\cite{Kondor2002} to provide a flexible representation of the spatial constraints, which can be combined with the landscape feature similarity matrix to produce a similarity graph for applying spectral clustering. 
In this paper, we extend the formulation to create hierarchical regions, where fine-scaled regions can be nested within broad-scale regions. Creating such nested regions is extremely useful for many applications because hierarchical structure is often held up as a fundamental feature of both the natural world and complex systems (as reviewed in \cite{WuJ2013}). In fact, the world’s biomes and ecological regions have often been delineated in a nested hierarchical structure \cite{Bailey2009,AbellRobin2008}. Constrained versions of hierarchical clustering techniques \cite{murtagh1985} such as single-link \cite{numericaleco}, complete-link \cite{rao2003}, UPGMA \cite{Kreft2010,Sandin1999}, and Ward's method \cite{Iyigun2013,Wolock2004} have often been used to create such nested regions. However, as will be shown in this study, the regions generated by such methods tend to be highly imbalanced in terms of their sizes, and thus, are not as suitable for many applications, including resource planning and management.  

We use a recursive bisection approach to extend our formulation in \cite{Yuan2015} to hierarchical clustering. Our top-down approach for creating nested regions is different from the bottom-up approach commonly used by existing methods~\cite{Iyigun2013,rao2003,Kreft2010,Sandin1999,Wolock2004}. Using three criteria for region evaluation---landscape homogeneity, region contiguity, and region size---our experimental results suggest that the proposed framework outperforms three other constrained hierarchical clustering methods in 2 out of the 3 criteria. For example, it consistently produces regions that are more homogeneous and balanced in region size compared to the spatially constrained complete-link \cite{rao2003} and UPGMA \cite{Kreft2010,Sandin1999} algorithms. Our proposed algorithm also outperforms the constrained version of Ward's method \cite{Wolock2004} in terms of producing regions that are spatially contiguous and approximately uniform in size. Finally, although the spatially constrained single link method \cite{numericaleco} is also capable of producing regions that are homogeneous and contiguous, it tends to create one or two very large regions that cover the majority of the landscape area. An ad-hoc parameter for maximum region size is needed by the spatially constrained single link method to prevent the formation of such large regions \cite{recchia2010}. Tuning this parameter is cumbersome as it must be done at every level of the hierarchy since the maximum region size depends on the number of regions. Our proposed hierarchical method does not have such a problem because its objective function, which is based on the normalized cut criterion \cite{Shi97normalizedcuts} used in spectral clustering, is inherently biased towards producing more uniformly-sized clusters.

The remainder of this paper is organized as follows. Section~\ref{sec:relatedwork} reviews previous work on the development of regionalization frameworks, constrained clustering, and hierarchical clustering methods. Section \ref{sec:prelim} formalizes the region delineation problem and presents an overview of spectral clustering. Section
\ref{sec:spatialconstainedsp} describes the different ways in which spatial constraints can be incorporated into the spectral clustering framework. It also presents the partitional and hierarchical implementations of our proposed spatially constrained spectral clustering framework. 
Section \ref{sec:application} describes the application of spatially constrained spectral clustering algorithms to the region delineation problem.  Section \ref{sec:con} concludes with a summary of the results of this study. 

\section{Related Work}\label{sec:relatedwork}
Region delineation has traditionally been studied as a spatial clustering~\cite{han2001} problem.  Duque et al.~\cite{Duque2007} classified the existing data-driven approaches into two categories. The first category does not require explicit representation and incorporation of spatial constraints into the clustering procedure. Instead, the constraints are satisfied by post-processing the clusters or optimizing other related criteria. For example, Openshaw~\cite{Openshaw1973} applied a conventional clustering method followed by a cluster refinement step to split clusters that contained geographically disconnected patches. %Weaver and Hess~\cite{Weaver1963} introduced a method for designing political districts by maximizing the compactness of the regions. 
The second category of methods explicitly incorporates spatial constraints into the clustering algorithm \cite{Duque2007}. Examples of such methods include adapted hierarchical clustering, exact optimization methods, and graph theory based methods.
%Zoltners and Sinha~\cite{Zoltners1983} proposed an method that solve the exact optimization problem. ~\cite{Duque2004}~\cite{Duque2006}, Heuristic models and mixed heuristic models.
This second category also encompasses the constrained clustering methods \cite{bacao2004,Wagstaff01constrainedk-means,Davidson05agglomerativehierarchical} developed in the fields of data mining and machine learning. %Constrained clustering algorithms 
Constrained clustering \cite{basu2008,Wagstaff01constrainedk-means} is a semi-supervised learning approach that uses the domain information provided by users to improve clustering results. The domain information is typically provided as must-link (ML) and cannot-link (CL) constraints to be satisfied by the clustering solution. ML constraints restrict the pairs of data points that must be assigned to the same cluster, whereas CL constraints specify the pairs of points that must be assigned to different clusters. For example, Kamvar et al.~\cite{Kamvar03spectrallearning} uses the ML and CL constraints to define the affinity matrix of the data. %Xu et al.~\cite{Xu05constrainedspectral} extended the approach in \cite{Kamvar03spectrallearning} by considering a random walk matrix instead of normalized graph Laplacian matrix. %Lu and Carreira-Perpi{\~{n}}{\'{a}}n~\cite{LuC08} proposed a method to propagate  constraints via Gaussian process. Ji and Xu~\cite{Ji:2006} introduced L2 regularization to penalize the violations of constraints in the clustering solution. 
Shi et al. \cite{ShiFY2010} proposed a constrained co-clustering method that considers both the similarity of features as well as the ML and CL constraints. All of these methods were designed to manipulate the graph Laplacian matrix using the domain constraints available. There has also been growing interest in developing constrained-based approaches for spectral clustering~\cite{Kamvar03spectrallearning,BoleyK13,ShiFY2010,Wang2010,craddock2012whole}.
%Alternatively, the constrained spectral clustering method can be designed to manipulate the feasible solution space of its optimization problem. 
For example, De Bie et al.~\cite{BieSM04} developed an approach that restricts the eigenspace for which the cluster membership vector is projected. 
%Coleman et al.~\cite{Coleman:2008} extended the approach proposed in ~\cite{BieSM04} to handle the situation in which some of constraints are potentially inconsistent with each other. Finally, 
Wang and Davison~\cite{Wang2010} proposed a constrained spectral clustering method that considers real-valued constraints and imposed a threshold on the minimum amount of constraints that must be satisfied by the feasible solution.  However, none of these constrained spectral clustering methods were designed for the region delineation problem. The framework presented in our previous paper \cite{Yuan2015,Kendra2017} employs a Hadamard product to combine the feature similarity matrix with spatial contiguity constraints, which is similar to the approach used in Craddock et al.~\cite{craddock2012whole} for generating an ROI atlas of the human brain using fMRI data. However, unlike the approach used in \cite{craddock2012whole}, we consider a truncated exponential kernel to relax the spatial neighborhood constraints and perform extensive experiments comparing the framework to various constrained spectral clustering algorithms.

Current constrained spectral clustering algorithms have also focused primarily on partitional clustering. %Although these methods are designed to find an optimal solution that minimizes the number of violated constraints, 
They require the number of clusters to be specified \textit{a priori}. %There are many ways to determine the number of clusters. For example, we can compute the sum of square within(SSW) cluster for different number of clusters and compare the same metric with random assignment. A heuristic way is to identify the 'elbow' of the slope or the gap of the two curves. Other metrics such as gap statistics\cite{Tibshirani00estimatingthe} is defined to estimate the number of clusters. However, for real world dataset such 'elbow' is difficult to identify and there is no unique rule to determine the number of clusters. 
In contrast, hierarchical methods generate a nested set of clustering for every possible number of clusters. The hierarchy of clusters, also known as a dendrogram, can be created either in a top-down (i.e., divisive hierarchical clustering) or bottom-up (i.e., agglomerative hierarchical clustering) fashion \cite{Tan:2005,Jain:1999,Jain:1988:ACD:46712}. %\SC{Is this what Kendra is talking about in the abstract and intro comments?  it seems like either way, we are still combining hu12 units into clusters, but that there are td or bu ways to do that?  or maybe I am just out of the loop on this.  either way, it might be good to give a better summary of this idea earliner in the intro or abstract to set up this more detailed explanation here?} \PN{Yes. This is what Kendra refers to.}
Some of the widely used agglomerative hierarchical clustering algorithms include single link~\cite{Iyigun2013}, complete link~\cite{rao2003}, group average (UPGMA)~\cite{Kreft2010,Sandin1999}, and Ward's method \cite{Jain:1999,Wolock2004} whereas examples of divisive hierarchical clustering algorithms include minimum spanning tree \cite{grygorash2006} and bisecting k-means \cite{savaresi2004}. Current approaches for creating nested regions are mostly based on different variations of agglomerative hierarchical clustering. Each of these variations has its own strengths and limitations \cite{murtagh1985}. For example, the single-link method can identify irregular shaped clusters but is highly sensitive to noise \cite{Tan:2005}. In contrast, the Ward's method can minimize the cluster variance but is susceptible to the inversion problem \cite{murtagh1985}. Unfortunately, many of these agglomerative methods can produce highly imbalanced sizes of regions, which is not desirable for many applications \cite{banerjee2006}.

\section{Preliminaries}
\label{sec:prelim}
This section formalizes region delineation as a constrained clustering problem and presents a brief overview of spectral clustering and its constrained-based methods. 

\subsection{Region Delineation as Constrained Clustering Problem}

Consider a data set $\mathcal{D} = \{(\mathbf{x}_i,\mathbf{s}_i)\}_{i=1}^N$, where $\mathbf{x}_i \in \mathbb{R}^d$ is a $d$-dimensional vector of landscape features associated with the geo-referenced spatial unit $\mathbf{s}_i \in \mathbb{R}^2$. Let $\mathcal{R} = \{1, 2, \cdots, k\}$ denote the set of region identifiers, where $k$ is the  number of regions, and $\mathcal{C} = \{(\mathbf{s}_i,\mathbf{s}_j,\mathbf{C}_{ij})\}$ denote the set of spatial constraints. For region delineation, we consider only ML constraints and represent them using a constraint matrix $\mathbf{C}$ defined as follows:
\begin{equation}
\mathbf{C}_{ij} = \begin{cases}
1 & \textrm{if $\mathbf{s}_i$ and $\mathbf{s}_j$ are spatially adjacent,} \\
0 & \textrm{otherwise.}
\end{cases}
\label{eqn:ml_const}
\end{equation}
The goal of region delineation is to learn a partition function $\mathcal{V}$ that maps each spatial unit $\mathbf{s}_i$ to its corresponding region identifier $r_i \in \mathcal{R}$ in such a way that (1) maximizes the similarity between the spatial units in each region and (2) minimizes the constraint violations in the set $\mathcal{C}$.

\subsection{Spectral Clustering}
\label{sec:spectral}

Spectral clustering is a class of partitional clustering algorithms that relies on the eigen-decomposition of an input affinity (similarity) matrix $\mathbf{S}$ to determine the underlying clusters of the data set. %Their empirical performance is generally superior to traditional clustering methods such as k-means, which are typically limited to finding compact, globular-shaped clusters \cite{Tan:2005}. The spectral clustering algorithm is also very flexible and can accommodate any type of similarity functions. Finally, it is equivalent to a relaxed version of the mincut graph partitioning problem~\cite{Shi97normalizedcuts}, which makes it an attractive approach for constrained clustering because the ML and CL constraints can also be represented as a constraint graph.
Let $\{\mathbf{x}_1,\mathbf{x}_2,\cdots, \mathbf{x}_N\}$ be a set of points to be clustered. To apply spectral clustering, we first compute an affinity matrix $\mathbf{S}$ between every pair of data points. The affinity matrix is used to construct an undirected weighted graph $\mathcal{G} = (V,E)$, where $V$ is the set of vertices (one for each data point) and $E$ is the set of edges between pairs of vertices. The weight of each edge is given by the affinity between the corresponding pair of data points. The Laplacian matrix of the graph is defined as $\mathbf{L} = \mathbf{D} - \mathbf{S}$, where $\mathbf{D}$ is a diagonal matrix whose diagonal elements correspond to $\mathbf{D}_{ii} = \sum_j \mathbf{S}_{ij}$. The goal of spectral clustering is to create a set of partitions on the graph $\mathcal{G}$ in such a way that minimizes the graph cut while maintaining a balanced size of the cluster partitions \cite{Luxburg:2007sp}.

%The graph Laplacian matrix has the following properties~\cite{Luxburg:2007sp}:
%\begin{enumerate}
%  \item $\mathbf{L}$ is a symmetric and positive semi-definite matrix.
%  \item The eigenvalues of $\mathbf{L}$ are non-negative and real-valued, $\lambda_n\ge\lambda_{n-1}\ge...\ge\lambda_1=0.$
%  \item The eigenvector that corresponds to eigenvalue 0 is a constant vector, whose elements are equal to 1 (if the graph has only one connected component).
%\end{enumerate}
The spectral clustering solution can be found by solving the following optimization problem~\cite{Luxburg:2007sp}:
\begin{equation}
\arg\min_{r} r^T\mathbf{L}r \ \textrm{s.t.} \ r^T\mathbf{D}r = \sum_i \mathbf{D}_{ii}, \ \mathds{1}^T\mathbf{D}r = \mathbf{0} 
\label{eqn:spectralcluster}
\end{equation}
where $\mathds{1}$ and $\mathbf{0}$ are vectors whose elements are all 1s and 0s, respectively. The solution for $r$ is obtained by solving the following generalized eigenvalue problem: 
%\begin{equation}
$\mathbf{L}r = \lambda \mathbf{D}r.$
%\label{eqn:gen_eig}
%\end{equation}
To obtain $k$ clusters, we first extract the top $k$ generalized eigenvectors and apply a standard clustering algorithm such as k-means to the data matrix generated from the eigenvectors.  

\subsection{Constrained Spectral Clustering}
\label{sec:cons_spec}
Current methods for incorporating constraints into spectral clustering algorithms can be divided into two categories. The first  category encompasses methods that directly alter the graph Laplacian matrix, e.g., 
%For example, Kamvar et al.~\cite{Kamvar03spectrallearning} employed a Gaussian kernel as their similarity matrix and considered a binary constraint matrix where ML constraints are assigned as 1 and CL constraints are designated as 0. Such an approach is inapplicable when some pairs of points have no constraints. Kawale and Boley~\cite{BoleyK13} propose to add L1 regularization that penalize the constraint violation to the graph cut objective function. Then the corresponding convex subproblem is solved. Due to the design of the constraint, the method only works for two class clustering problem. Therefore, in this section, we focus on three formulation of constraint incorporation.
by applying a weighted sum between the feature similarity matrix $\mathbf{S}$ and the constraint matrix $\mathbf{C}$ given in Equation \eqref{eqn:ml_const}:
\begin{equation}
\textrm{Weighted sum:} \ \ \ \mathbf{S}^{\textrm{total}}(\delta) = (1-\delta) \mathbf{S} + \delta \mathbf{C}, 
\label{eqn:weighted}
\end{equation}
$\delta \in [0,1]$ is a parameter that controls the trade-off between maximizing cluster homogeneity and preserving the  constraints of the data. When $\delta$ approaches zero, the clustering solution is more biased towards maximizing the feature similarity whereas when $\delta$ approaches one, it is more biased towards preserving the constraints. 

Let $\mathbf{D}$ and $\mathbf{D}^{(c)}$ be the diagonal matrices constructed from the feature similarity matrix ($\mathbf{S}$) and constraint matrix ($\mathbf{C}$) in the following way:
$$\mathbf{D}_{ii} = \sum_{j} \mathbf{S}_{ij}, \ \ \ \mathbf{D}^{(c)}_{ii} = \sum_j \mathbf{C}_{ij}.$$ Using Equation \eqref{eqn:weighted}, it can be shown that the modified graph Laplacian is given by a convex combination of the graph Laplacian for the feature similarity matrix and the graph Laplacian for the constraint matrix, i.e.,
\begin{eqnarray}
%\mathbf{D}^{\textrm{total}}_{ii} &=& \sum_j  \mathbf{S}^{\textrm{total}}_{ij}(\delta) = (1 - \delta) \mathbf{D}_{ii} + \delta \mathbf{D}^{(c)}_{ii} \nonumber \\
\mathbf{L}^{\textrm{total}} &=& \mathbf{D}^{\textrm{total}} - \mathbf{S}^{\textrm{total}} \nonumber \\
&=& (1 - \delta)(\mathbf{D} - \mathbf{S}) + \delta (\mathbf{D}_c - \mathbf{C}) 
%&=& (1 - \delta)\mathbf{L} + \delta \mathbf{L}_c 
\end{eqnarray}

The weighted sum approach described above is a special case of the spectral constraint modeling (SCM) algorithm proposed by Shi et al. \cite{ShiFY2010}. %for co-clustering problems. 
The altered graph Laplacian can be substituted into Equation \eqref{eqn:spectralcluster}, which in turn, allows us to apply existing spectral clustering algorithm to identify the regions.
\begin{eqnarray} 
\label{eqn:scm}
\textrm{SCM:} \ \ \ \ && \operatorname*{arg\,min}_{r\in\mathbb{R}^N} r^T\mathbf{L}^{\textrm{total}}r \\
\textrm{s.t.} && r^T\mathbf{D}^{\textrm{total}}r = \sum_i \mathbf{D}^{\textrm{total}}_{ii},\ \mathds{1}^T\mathbf{D}^{\textrm{total}}r = \mathbf{0}. \nonumber
\end{eqnarray}

%\[ \operatorname*{min}_{v\in\mathbb{R}^N} v^TLv + \delta v^TL_cv\] \[s.t.\ v^T1 = 0, v^Tv = 1 \]Where $\delta $  is a user defined parameter.

The second category of approaches for incorporating domain constraints is to alter the feasible solution set of the spectral clustering algorithm. For example, Wang and Davidson \cite{Wang2010} proposed the CSP algorithm, which optimizes the following objective function. 
\begin{eqnarray} 
\label{eqn:CSP}
\textrm{CSP:} \ \ \ \ && \operatorname*{arg\,min}_{r\in \mathbb{R}^N} r^T\bar{\mathbf{L}}r \\ 
\textrm{s.t.} && r^T\bar{\mathbf{C}}r\ge\alpha,\ r^Tr = vol(\mathcal{G}),\ r\ne \mathbf{D}^{1/2} 1, \nonumber
\end{eqnarray}
where $\bar{\mathbf{L}} = \mathbf{D}^{-1/2}\mathbf{L}\mathbf{D}^{-1/2}$ and $\bar{\mathbf{C}} = \mathbf{D}_c^{-1/2}\mathbf{C}\mathbf{D}_c^{-1/2}$ are the normalized graph Laplacian and normalized constraint matrix, respectively. The threshold $\alpha$ gives a lower bound on the amount of constraints in $\mathbf{C}$ that must be satisfied by the clustering solution. Instead of setting the parameter for $\alpha$, Wang and Davison~\cite{Wang2010} requires users to specify a related parameter $\beta$, which was shown to be a lower bound for $\alpha$.

%As $\lambda_{max}(\bar{C})>0$, we define a tuning parameter $\delta = \frac{\beta}{\lambda_{max}vol}, \delta \in (-\infty, 1).$

%\[ \operatorname*{arg\,min}_{v\in\mathbb{R}^N} v^T\bar{L}v\] \[s.t.\ v^T\bar{Q}v\ge\alpha,\ v^Tv = vol(G),\ v\ne D^{1/2} 1 \] where the cost of the cut $v^T\bar{L}v$ is minimized and $v^T\bar{Q}v\ge\alpha$ controls the lower bound of how well the user specified constraint being satisfied.

\section{Spatially Constrained Spectral Clustering} 
\label{sec:spatialconstainedsp}

In this section, we describe the various ways to represent spatial contiguity constraints and to incorporate them into the spectral clustering framework. 

%\subsection{Constrained Spectral Clustering}
%Traditional clustering is unsupervised\cite{Jain:1999}. In real world, however,  clustering based one\subsection{Spectral Clustering}

\subsection{Kernel Representation of Spatial Contiguity Constraints}

For constrained spectral clustering, we can define a corresponding constraint graph $\mathcal{G}_C = (V,E_C)$, where $V$ is the set of data points and $E_C$ is the set of edges whose weights are defined as follows:
\begin{equation}
E_{ij} = \left\{
  \begin{array}{ll}
    1, & \hbox{$(v_i,v_j)$ is a ML edge;} \\
    -1, & \hbox{$(v_i,v_j)$ is a CL edge;} \\
    0, & \hbox{otherwise.}
  \end{array}
\right.
\label{eqn:constraints}
\end{equation}

For region delineation, the vertices of the constraint graph correspond to the set of spatial units to be clustered, while the ML edges correspond to pairs of spatial units that are adjacent to each other. It is also possible to define a CL edge between every pair of spatial units that are either located too far away from each other or are obstructed by certain barriers (e.g., large  bodies of water) that make them unreasonable for assignment to the same region. However, since the number of CL edges tends to grow almost quadratically with increasing number of points, this severely affects the runtime of spectral clustering algorithm. Furthermore, the ML edges are often sufficient to provide guidance on how to form spatially contiguous regions. For these reasons, we consider constraint graphs that have ML edges only in this paper. Let $\mathbf{C}$ denote the adjacency matrix representation of the edge set $E_C$. 

\begin{figure}[t!]
\centering
    \includegraphics[width = 5cm]{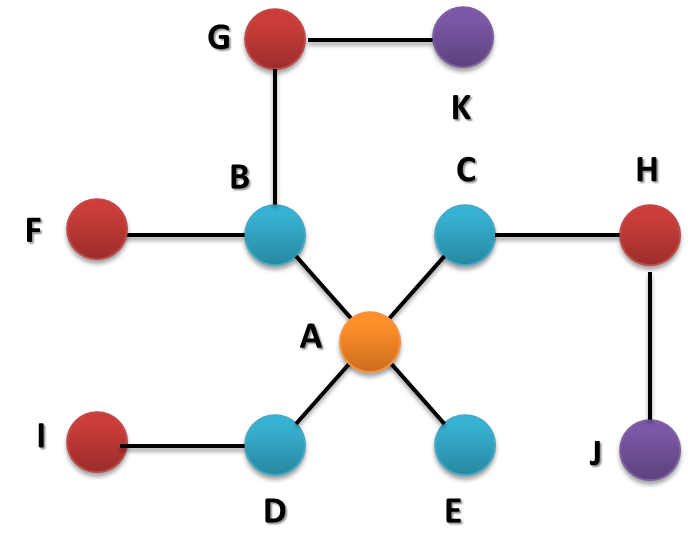}%[trim = 135mm 42mm 5mm 25mm, clip, width=50mm]{toyexample.png}
%    \captionlistentry[table]{A table beside a figure}
%    \captionsetup{labelformat=andtable}
    \caption{An illustration of spatial contiguity constraint}
    \label{fig:spatial_contiguity}
  \end{figure}

A constrained spectral clustering algorithm is designed to produce solutions that are consistent with the constraints imposed by $\mathcal{G}_C$. Unfortunately, for region delineation, it may not be sufficient to use the adjacency information between neighboring spatial units to control the trade-off between spatial contiguity and landscape homogeneity of the regions. To improve its flexibility, we introduce a \textit{spatially constrained kernel matrix}, $\mathbf{S}_c$. The simplest form of the kernel would be a linear kernel, which is defined as follows:
\begin{equation}
\textrm{Linear Kernel:} \hspace{1.5cm} \mathbf{S}_c^{\textrm{linear}} = \mathbf{C} 
\label{eqn:lin_kernel}
\end{equation}
More generally, we can define an exponential kernel~\cite{Kondor2002} on the adjacency matrix $\mathbf{C}$ as follows.
\begin{eqnarray}
\textrm{Exponential Kernel:} \hspace{5cm} \nonumber \\
\mathbf{S}_c^{\textrm{exp}} = e^\mathbf{C}  = \mathds{I} +\mathbf{C} + \frac{1}{2!}\mathbf{C}^2+\frac{1}{3!}\mathbf{C}^3+\cdots = \sum_{k = 0}^{\infty}\frac{\mathbf{C}^k}{k!} 
\label{eqn:exp_kernel}
\end{eqnarray}
where $\mathds{I}$ is the identity matrix. Since we consider only ML constraints, the $k$-th power of the adjacency matrix $\mathbf{C}$ represents the number of ML paths of length $k$ that exist between every pair of vertices. An ML path between vertices $(v_i,v_j)$ refers to a sequence of ML edges $e_1, e_2, \cdots, e_m$ such that the initial vertex of $e_1$ is $v_i$ and the terminal vertex of $e_m$ is $v_j$. It can be shown that $\mathbf{S}_c^{\textrm{exp}}$ is a symmetric, positive semi-definite matrix, and thus, is a valid kernel~\cite{Kondor2002}. Furthermore, as the diameter of the constraint graph is finite, we also consider a truncated version of the exponential kernel:
\begin{equation}
\textrm{Truncated Exponential}: \ \ 
\mathbf{S}_c^{\textrm{trunc}}(\delta) \equiv \sum_{k = 0}^{\delta}\frac{\mathbf{C}^k}{k!} 
\label{eqn:exp_kernel2}
\end{equation}
where the parameter $\delta$ controls the ML neighborhood size of a vertex. The ML neighborhood specifies the set of vertices that should be in the same region as the vertex under consideration. 
%When $\delta = 1$, the truncated exponential kernel reduces to a linear kernel.
As an example, consider the graph shown in Figure \ref{fig:spatial_contiguity}. When $\delta=1$, the ML neighborhood for vertex A corresponds to its immediate neighbors, B, C, D and E. When $\delta = 2$, the ML neighborhood of vertex A is expanded to include vertices that are located within a path of length 2 or less from A, i.e., B, C, D, E, F, G, H and I. When $\delta = 3$, the ML neighborhood for vertex A includes all of the vertices in the graph. Note that each term in the summation given in Equation \eqref{eqn:exp_kernel} is normalized by the path length; therefore, a vertex that is located further away from a given vertex has less influence as compared to a nearer vertex.

Finally, the truncated exponential kernel matrix can be binarized so that it can be interpreted as an adjacency matrix for an expanded constraint graph, whose ML neighborhood size is given by the parameter $\delta$.
\begin{eqnarray}
\textrm{Binarized Truncated Exponential Kernel}: \hspace{2cm} \nonumber \\ 
\mathbf{S}_c^{\textrm{bin}}(\delta) \equiv \mathbf{I} \bigg[\sum_{k = 0}^{\delta}{\mathbf{C}^k > 0}\bigg] \hspace{2.5cm}
\label{eqn:exp_kernel3}
\end{eqnarray}
where $\mathbf{I}[\cdot]$ is an indicator function whose value is equal to 1 if its argument is true and 0 otherwise. Both the truncated and binarized truncated exponential kernels allow us to vary the degree to which the original constraint graph should be satisfied. As $\delta$ increases, the constraint satisfaction becomes more relaxed. Ultimately, when $\delta$ is greater than or equal to the diameter of the graph, $\mathbf{S}_c^{\textrm{bin}}$ reduces to a matrix of all 1s, which is equivalent to ignoring the spatial contiguity constraints. 

\subsection{Hadamard Product Graph Laplacian}

We now describe our approach for incorporating the spatially constrained kernel matrix $\mathbf{S}_c$ into the spectral clustering formulation. Instead of using the weighted sum approach given in Equation \eqref{eqn:weighted}, we consider a Hadamard product approach to combine $\mathbf{S}_c$ with the feature similarity matrix 
$\mathbf{S}$:
\begin{equation}
\textrm{Hadamard Product:} \ \ \ \ \ \mathbf{S}^{\textrm{total}}(\delta) = \mathbf{S} \circ \mathbf{S}_c(\delta), \label{eqn:hadamard}
\end{equation}
where $\mathbf{S}_c(\delta)$ corresponds to either the truncated exponential kernel (Equation  \eqref{eqn:exp_kernel2}) or the binarized truncated exponential kernel (Equation \eqref{eqn:exp_kernel3}). 

There are several advantages to using a Hadamard product approach to combine the matrices. First, unlike the weighted sum approach, it discourages spatial units that are located far away from each other from being assigned to the same cluster even though their feature similarity is high. Second, it produces a sparser kernel matrix, which is advantageous for large-scale graph analysis. %\cite{HeZHK11}\cite{Yan2009siam}. 
Finally, it gives more flexibility to the users to specify the level of constraints that must be preserved by tuning the parameter $\delta$, which controls the ML neighborhood size of the constraint graph. %Its value ranges between 1 and the diameter of the constraint graph $\mathcal{G}_c$. 

Let $\mathbf{D}^{\textrm{total}}_{ii} = \sum_j [\mathbf{S} \circ  \mathbf{S}_{(c)}(\delta)]_{ij}$  be elements of a diagonal matrix computed from  $\mathbf{S}^{\textrm{total}}$. The Hadamard product graph Laplacian is given by $\mathbf{L}^{\textrm{total}} = \mathbf{D}^{\textrm{total}} - \mathbf{S} \circ \mathbf{S}_c(\delta)$. The modified graph Laplacian can be substituted into Equation \eqref{eqn:spectralcluster} and solved using the generalized eigenvalue approach to identify the regions.

\subsection{Partitional Spatially-Constrained Spectral Clustering Algorithm}
\label{sec:ssc_framework}

%we propose our spatial clustering that achieve the goal of maintaining the spatial contiguity. Our framework illustrate an trade off between satisfaction of constraint and cluster quality. A summary of the framework is presented in Algorithm \ref{alg:1}.

\begin{algorithm}[t!]
    %\begin{tabular}{|p{8cm}|}
\textbf{Input:}\\
$\mathcal{D} = \{(\mathbf{x}_1,\mathbf{s}_1), (\mathbf{x}_2,\mathbf{s}_2),..., (\mathbf{x}_N,\mathbf{s}_N)\}$ \\
$\mathbf{C} \in R^{N\times N}$: spatial constraint matrix.\\
%$\mathbf{C} \in R^{n\times n}$: spatial constraint matrix. \\
$k$: number of clusters.\\
$\delta$: neighborhood size.\\
\textbf{Output:}\\
$\mathcal{R}$ = $\{R_1, R_2,..., R_{k}\}$ (set of regions). \\

\textbf{1.} Create similarity matrix $\mathbf{S}$ from $\{\mathbf{x}_1, \mathbf{x}_2, \cdots, \mathbf{x}_N\}$.\\
\textbf{2.} Compute the spatially constrained kernel matrix, $\mathbf{S}_c(\delta)$. \\
\textbf{3.} Compute the combined kernel $\mathbf{S}^{\textrm{total}}$ based on $\mathbf{S}$ and $\mathbf{S}_c$.\\ 
\textbf{4.} Compute  $\mathbf{D}^{\textrm{total}}$ and  $\mathbf{L}^{\textrm{total}}$.\\
\textbf{5.} Solve the generalized eigenvalue problem $\mathbf{L}^{\textrm{total}} r = \lambda \mathbf{D}^{\textrm{total}} r$. Create matrix $\mathbf{X}_r = [r_1 r_2 \cdots r_k]$ from the top-k eigenvectors.\\
\textbf{6.} $\mathcal{R}$ $\leftarrow$ \texttt{k-means($\mathbf{X}_r$,$k$)}\\

\normalsize \caption{Partitional Spatially-Constrained Spectral Clustering} \label{alg:1}
\end{algorithm}

Algorithm~\ref{alg:1} presents a high-level overview of our partitional clustering approach. First, a feature similarity matrix is created by applying the Gaussian radial basis function kernel, 
$k(\mathbf{x}_i,\mathbf{x}_j) = \exp(-\frac{||\mathbf{x}_i-\mathbf{x}_j||^2}{2\sigma^2})$ to the feature set of the spatial units. The spatially constrained kernel matrix $\mathbf{S}_c$ is then computed from the constraint matrix $\mathbf{C}$, where $\mathbf{C}_{ij} = 1$ if $(\mathbf{s}_i, \mathbf{s}_j)$ is a ML edge and 0 otherwise. Note that if the truncated exponential kernel is used to represent the spatially constrained kernel matrix, we termed the approach as a spatially-constrained spectral clustering (SSC) algorithm. However, if the binarized truncated exponential kernel is used, the approach is known as a binarized spatially-constrained spectral clustering (BSSC). 

Once the combined graph Laplacian, $\mathbf {L}^{\textrm{total}}$ is found, we extracted the first k %of its 
eigenvectors as the low rank approximation of the combined kernel matrices.  We then applied %simple 
k-means clustering to partition the data into its respective regions. Note that the partitional clustering framework shown in Algorithm \ref{alg:1} is also applicable to the SCM and CSP algorithms, by setting their corresponding graph Laplacian, $\mathbf{L}^{\textrm{total}}$ and diagonal matrix, $\mathbf{D}^{\textrm{total}}$. %The algorithm first compute the unnormalized graph Laplacian matrix L. Then it computes the first k generalized eigenvectors of the function $Lu = \lambda Du$. Finally, apply k-means on the first k eigenvectors. The output of the algorithm is a vector indicate the cluster assignment for each data point.
The computational complexity of the
%\SY{For a data set containing n data points, The complexity of constructing both $\mathbf{S}$ and $\mathbf{S_c}$ is $\mathbf{O}(n^2)$. In general, the eigen-decomposition will cost $\mathbf{O}(n^3)$. Therefore, the complexity of
spatially constrained spectral clustering is equivalent to the standard spectral clustering algorithm, which is $O(N^3)$. 

\subsection{Hierarchical Spatially-Constrained Spectral Clustering Algorithm}
\label{sec:hbssc}

%As previously noted, nested regions are important for many applications as they enable properties of a geographical system to be investigated at varying spatial scales. 
The formulation described in the previous section can be extended to  hierarchical clustering by using a recursive bisection approach. Specifically, the algorithm will iteratively identify the least homogeneous region to be split into two smaller subregions until every subregion contains only a single spatial unit.
%Previously, we introduced two approaches SSC and BSSC which are designed to split data into k clusters. In this section, we provide a divisive hierarchical version of the spatially constrained spectral clustering (HSSC). 

The pseudocode of the proposed algorithm is shown in Algorithm \ref{alg:2}. First, the feature similarity matrix $\mathbf{S}$ is computed using the Gaussian RBF kernel function. Next, the spatial constraint matrix $\mathbf{S}_c(\delta)$ is created using Equation \ref{eqn:exp_kernel3}. The algorithm will then compute the combined kernel $\mathbf{S}^{\textrm{total}}$ and its corresponding graph Laplacian matrix $\mathbf{L}^{\textrm{total}}$, similar to the approach described in Section \ref{sec:cons_spec}. The algorithm initially assigns all the data points to a single cluster. It then recursively partitions the data until $k$ clusters are obtained, as shown in lines 6a-6d in Algorithm \ref{alg:2}. Let $R_{k-1}$ be the set of clusters found after $k-1$ iterations. On line 6a, the algorithm chooses the cluster $\mathcal{C}^k \in R_{k-1}$ with the worst sum of square within errors (SSW) to be split into two smaller clusters, $C_1$ and $C_2$ (line 6c). One advantage of using our top-down recursive partitioning approach is that neither the feature similarity nor the spatial constraint matrix have to be updated at each iteration unlike the bottom-up hierarchical clustering, which requires us to re-compute the modified feature similarity and constraint matrices each time a pair of clusters is merged. 
% It then applies spectral clustering on $\mathbf{L^{total}}$ to partition the data set into 2 disjoint clusters. \SY{We use the sum of square within clusters(SSW) as a evaluation metric of the quality of the clusters. The higher SSW, the lower the homogeneity of the cluster. We compute SSW for each cluster} and the cluster with highest SSW is chosen to be split in the next iteration. This process is repeated until all number of clusters is obtained. 

%The advantages of using the hierarchical spatially constrained spectral clustering are as follows. First, the resulted cluster assignment can be any number of clusters. Secondly, the output is a nested cluster assignment. Thirdly, this partition more likely to produce clusters of a balanced cluster size.

\begin{algorithm}[t!]
\textbf{Input:}\\
$\mathcal{D} = \{(\mathbf{x}_1,\mathbf{s}_1), (\mathbf{x}_2,\mathbf{s}_2),..., (\mathbf{x}_N,\mathbf{s}_N)\}$ \\
$\mathbf{C} \in R^{N\times N}$: spatial constraint matrix.\\
$\delta$: neighborhood size.\\
\textbf{Output:}\\
$\mathcal{R}$ = $\{R_1, R_2,..., R_{k}\}$ (set of regions). \\

\textbf{1.} Create similarity matrix $\mathbf{S}$ from $\{\mathbf{x}_1, \mathbf{x}_2, \cdots, \mathbf{x}_N\}$.\\
\textbf{2.} Compute the spatially constrained kernel matrix, $\mathbf{S}_c(\delta)$. \\
\textbf{3.} Compute the combined kernel $\mathbf{S}^{\textrm{total}}$ based on $\mathbf{S}$ and $\mathbf{S}_c$.\\ 
\textbf{4.} Compute $\mathbf{D}^{\textrm{total}}$ and  $\mathbf{L}^{\textrm{total}}$.\\
\textbf{5.} Initialize $R_1$ as the cluster containing all $N$ spatial units. \\
\textbf{6.} \textbf{for} $k = 2$ to $N$ \textbf{do}\\
\textbf{6a.} $C^*$ = choose($R_{k-1}$) \\
\textbf{6b. }  $R_k$ $\leftarrow$ $R_{k-1} - C^*$ \\
\textbf{6c.} ($C_1$, $C_2$) $\leftarrow$ SSC($\mathcal{D}_{C^*}$, 2) \\
\textbf{6d.} $R_k$ $\leftarrow$ $R_{k-1} \cup \{C_1, C_2)$. \\
\normalsize \caption{Hierarchical Spatially-Constrained Spectral Clustering (HSSC)} \label{alg:2}
\end{algorithm}

\section{Application to Region Delineation} \label{sec:application}
%Duque et al.~\cite{Duque2007} provides a taxonomy of methods for regionalization where the spatial aggregation task has been categorized into two types. The first type with no explicit spatial constraint. Openshaw~\cite{Openshaw1973} proposed a two-stage algorithm that applies conventional clustering followed by a cluster refinement step to separate data points that are not geographically connected. The second type corresponds to methods with explicit spatial constraint. This includes exact optimization models~\cite{Zoltners1983}~\cite{Duque2004}~\cite{Duque2006}, Heuristic models and mixed heuristic models.

To evaluate the effectiveness of constrained spectral clustering for region delineation, we conducted a case study on a large-scale terrestrial ecology data set. %The performance of our proposed SSC and BSSC algorithms were compared against several baseline methods in terms of the spatial contiguity of their regions as well as homogeneity of the ecological features in the regions. 
The results of the case study are presented in this section. 

\subsection{Data set}

The constrained spectral clustering methods were assessed using geospatial data from the LAGOS$_\textrm{GEO}$ \cite{lagos2015,lagosne2017} database. The database %The data set used for our experiments~\cite{lagos2015} 
contains landscape characterization features measured at multiple spatial scales with a spatial extent that covers a land area spanning 17 U.S. states. The land area was divided into smaller hydrologic units (HUs), identified by their 12-Digit Hydrologic Unit Code~\cite{Huc1987}. % It includes the 12 digit HUCs from the Natural Resources Conservation Service(NRCS) watershed boundary dataset (downloaded July 2013) with at least ten percent of their land area within the 17 state LAGOS study region. 
Our goal was to develop a regionalization system for the landscape by aggregating the 20,257 HUs into coarser regions. %For clustering \SC{clustering sounds a bit strange. maybe demonstration, or "To demonstrate the clustering approach,"?} purposes, 
We selected 28 terrestrial landscape variables and performed experiments on three study areas---Michigan, Iowa, and Minnesota. %We selected 28 terrestrial landscape variables and divided the data into three study areas that are U.S. states: Michigan, Iowa, and Minnesota. 
When the values for a landscape variable was always zero, we removed that variable before applying the clustering methods. 
The number of HUs to be clustered in each study region, as well as number of landscape variables for each, are summarized in Table \ref{tab:dataset}.

\begin{table*}[h]
\centering
\caption{Summary statistics of the data set}
\label{tab:dataset}
\begin{tabular}{|c|c|c|c|c|} \hline
Study Area & \# HUs & \# landscape & \# PCA  & Diameter of  \\ 
& &variables&components&constraint graph\\
\hline \hline
Michigan & 1,796 & 17 & 10 & 41\\
Iowa & 1,605 & 19 & 12 & 43\\
Minnesota & 2,306 & 19 & 11&57 \\
%New Hampshire & 278 & 17 & 9\\ 
\hline
\end{tabular}
\end{table*}

% * <colli636@msu.edu> 2015-05-16T21:19:52.591Z:
%I changed all of the HUC to capitals to maintain consistency with the original definition
% 
%
% ^ <spencek1@gmail.com> 2015-05-17T15:48:22.328Z:
%
%  And, the HUC = hydro unit CODE, but they are actually called hydro units, so I changed most to HUs
%

The data set was further preprocessed before applying the constrained clustering algorithms. First, %missing values were interpolated by the average value of adjacent HUs. \SC{might be nice to indicate that there are few missing values - perhaps give a percentage or add a col to the table with number of hucs, etc} \PNT{No missing values for MN, MI, IA.} Each 
each variable was standardized to have a mean value of zero and variance of one. Since some of the landscape variables were highly correlated, we applied principal component analysis to reduce the number of features, keeping only the principal components that collectively explained at least 85\% of the total variance. %\SC{reword - it makes it sound like you only kept principal components that individually explained 85\%, which is clearly not right - instead perhaps "kept only the principal components that summed to explain at least 85\% of the total variance."?} 
The principal component scores were then used to calculate a feature similarity matrix for all pairs of HUs in each study area. The ML edges for the constraint graph were determined based on whether the polygons for two HUs were adjacent to each other. 

\subsection{Baseline Methods}\label{subsec:baseline}
%We consider several baseline methods for comparing the performance of our proposed methods. 
For partitional-based constrained clustering, we compared our algorithms, SSC and BSSC, against three competing baseline methods. The first baseline, called SCM~\cite{ShiFY2010}, uses a weighted sum approach (Equation \eqref{eqn:weighted}) to combine the feature similarity matrix $\mathbf{S}$ with the adjacency matrix $\mathbf{C}$ of the constraint graph. The algorithm has a parameter $\delta \in [0,1]$ that controls whether the clustering should favor homogeneity or spatial contiguity of the regions. When $\delta$ approaches 0, the algorithm is biased towards maximizing the similarity of features in the regions whereas when $\delta$ approaches 1, it is biased towards producing more contiguous regions.

The second baseline method, called  CSP~\cite{Wang2010}, uses the spatial constraints to restrict the feasible set of the clustering solution (Equation \eqref{eqn:CSP}). As noted in Section \ref{sec:cons_spec}, the algorithm has a parameter $\beta$ that gives a lower bound on the proportion of constraints that must be satisfied by the clustering solution. Furthermore, $\beta < \lambda_{\max} vol(\mathcal{G})$ to ensure the existence of a feasible solution \cite{Wang2010}. Instead of using $\beta$, we define an equivalent tuning parameter $\delta = \beta/ [\lambda_{max}vol(\mathcal{G})]$ so that its upper bound, which is equal to 1, is consistent with the upper bound for other algorithms evaluated in this study. %The algorithm will end when every data point is one single cluster or a maximum number of clusters is reached.  

%As $\lambda_{max}(\bar{C})>0$, we define a tuning parameter $\delta = \frac{\beta}{\lambda_{max}vol}, \delta \in (-\infty, 1).$

The third baseline is a spatially constrained clustering method proposed recently in the ecology literature by Miele et al.~\cite{miele2014}. It uses a stochastic model to represent nodes and links in a spatial ecological network. The cluster membership of each node is assumed to follow a multinomial distribution. Spatial constraints are introduced as a regularization penalty in the maximum likelihood estimation of the model parameters. The algorithm is implemented as part of the Geoclust R package. We denote the model-based method as MB in the remainder of this paper.  

For hierarchical clustering, we compare our proposed HSSC algorithm against the space-constrained clustering method described in \cite{numericaleco}. The method is similar to traditional agglomerative hierarchical clustering, except it applies a Hadamard product between the feature similarity matrix $\mathbf{S}$ with the spatial constraint matrix $\mathbf{S}_c$ to generate a combined similarity matrix $\mathbf{S}^{\textrm{total}}$. This is identical to the approach used in HSSC. %However, unlike HSSC, both $\mathbf{S}$ and $\mathbf{S}_c$ must be iteratively updated every time a pair of regions (clusters) is merged. The updated matrices are used to re-compute $\mathbf{S}^{\textrm{total}}$. 
The agglomerative clustering algorithm initially assigns each spatial unit to be in its own cluster (region). It then merges the two clusters with the highest similarity value in $\mathbf{S}^{\textrm{total}}$. Both the feature similarity matrix $\mathbf{S}$ and the spatial constraint matrix $\mathbf{S}_c$ are then updated accordingly. The update for $\mathbf{S}$ depends on how the similarity between two clusters is computed. Among the popular approaches that have been used to update $\mathbf{S}$ include single link~\cite{sneath1973numerical}, complete link~\cite{sorensen1948method}, group average (UPGMA)~\cite{sokal1958statistical}, and the Ward's method~\cite{ward1963hierarchical}. The adjacency matrix $\mathbf{C}$ is updated based on whether there is a path from any point in one cluster to any point in the other cluster and the constrained similarity matrix $\mathbf{S_c}$ is updated based on Equation \ref{eqn:exp_kernel2} with a predefined $\delta$. 

We implemented SCM, SSC, BSSC, HSSC and the spatially constrained agglomerative hierarchical clustering (single link, complete link, UPGMA, Ward's method) in Matlab. For CSP and MB, we downloaded their software from the links provided by the authors\footnote{CSP was obtained from \url{https://github.com/gnaixgnaw/CSP} whereas MB was downloaded from \url{http://lbbe.univ-lyon1.fr/Download-5012.html?lang=fr}.}. 

%It is a model based approach that assumes data and labels follow certain distributions. The labels can be retrieved by maximizing the complete likelihood function. The authors added a regularization term in the objective function so as to accounting for the spatial constraint. The formulation is the following:
%\[logL(Y,Z,\gamma)-\lambda \times pen(Z;L_x)\]
%The first term $logL(Y,Z,\gamma)$ is the complete data likelihood where $Y$ standing for the ecological features, $Z$ represent the label set, and $\gamma$ denote the parameter set governs the distribution of $Y$ and $Z$. The second term is to penalize the inconsistency in spatial information $L_x$ and class label $Z$. The optimization can be solved by EM algorithm.

\subsection{Evaluation Metrics}\label{sec:metrics}
We evaluated the performance of the algorithms based on three criteria: homogeneity, spatial contiguity, and region size. To determine whether the regions were ecologically homogeneous, we computed their within-cluster sum-of-square error (SSW) \cite{Tan:2005}:
\begin{equation}
\textrm{SSW} = \sum_{i=1}^k \sum_{x\in C_i}dist(\mu_i,x)^2 
\label{eqn:ssw}
\end{equation}
where $\mu_i$ is the centroid of the cluster $C_i$. The lower SSW is, the more homogeneous are the spatial units within the regions. 

%In supervised learning, algorithm can be evaluated by comparing the output with ground truth such as accuracy, precision and recall. When ground truth is unavailable, cluster cohesion and separation can be measure of cluster quality. Besides, we also needed to investigate how well the spatial constraint was satisfied. We used contiguity metrics introduced by Wu and Murray \cite{Wu2008}. %Further, we compare compare our results with previous ecology region formation work.
% * <colli636@msu.edu> 2015-05-16T21:28:14.828Z:
%ground truth or ground truthing?
% 
%
%\paragraph{Cluster cohesion and separation.} Cluster cohesion can be measured as sum of squares within cluster(SSW) which is defined as 
% Cluster separation can be defined in a similar way as the sum of squares between clusters:
%\begin{equation}
%SSB = \sum_{i=1}^{k}m_i\times dist(c_i,c)
%\end{equation}
%where $m_i$ is the number of points in cluster $C_i$ and c is the overall mean. Clusters with Higher SSB performs better separation.
% ^ <ptan@msu.edu> 2015-05-17T22:28:59.536Z:
%
%  I've removed the mentioning of ground truth (since we don't have any).
%

%\paragraph{Contiguity metric} Spatial contiguity is a 
%fundamental property for regionalization. For each cluster, the ideal case will contain only one connected components, that is there is no isolated areas or separated patches. 

%-------------Parameter tuning-------------------------------------------------

The second criteria assesses the spatial contiguity of the resulting  regions. We consider two metrics for this evaluation. The first metric computes the percentage of ML constraints preserved within the regions:
\begin{equation}
\textrm{PctML} = \frac{\textrm{\# ML edges within discovered regions}}{\textrm{Total \# of ML edges}}
\end{equation}
The second metric corresponds to a relative contiguity metric proposed in the ecology literature by Wu and Murray \cite{Wu2008}. The metric takes into consideration both the within patch contiguity ($\phi$) and between patch contiguity ($\nu$):
\begin{equation} 
c = \frac{\phi  + \nu}{\Omega}
\end{equation}
where
\begin{eqnarray}
\phi &=& \sum_{i = 1}^{k}(\frac{N_i(N_i-1)}{2}), \ \ 
\nu = \frac{1}{2}\sum_{i = 1}^k\sum_{j = 1,j\neq i }^k(\frac{N_i N_j}{l_{ij}^\gamma})\nonumber \\
\Omega &=& \frac{(\sum_{i = 1}^k N_i)(\sum_{i = 1}^k N_i -1 )}{2}\nonumber
\end{eqnarray}
In the preceding formula, $k$ is the number of regions and $N_i$ is the number of spatial units assigned to the $i$-th region. $l_{ij}$ denote the minimum spanning tree path length between regions $i$ and $j$ while $\gamma$ is a distance decay parameter. Since the metric is normalized by the total number of possible edges in a complete graph ($\Omega$), it ranges between 0 and 1. 

Although spatial contiguity is a desirable criterion, it may lead to highly imbalanced regions \cite{murtagh1985}. For example, an algorithm that creates one very large region along with many smaller but contiguous regions will likely have a high contiguity value. Previous studies~\cite{Ding02,banerjee2006} have shown the importance of maintaining a more balanced cluster sizes to ensure good clustering performance. Thus, %our third criteria measures the degree of balance of the cluster sizes. 
%The balance of cluster sizes is an important factor in order to achieve a good partition\cite{Ding02}. 
given a set of $k$ clusters %, $\mathcal{R}$ = $\{R_1, R_2,..., R_{k}\}$, 
with their corresponding cluster sizes, $n_1, n_2, ..., n_k$, we define a metric, $Cbalance$, based on the normalized geometric mean of the cluster sizes:  
\begin{eqnarray}
Cbalance = \frac{k}{N} \bigg[n_1\times n_2 \times ... \times n_k\bigg]^{\frac{1}{k}}, 
\end{eqnarray}
where $N$ is the total number of data points and $k$ is the number of clusters. The metric ranges from 0 to 1 and the larger the value, the more balanced are the cluster sizes.

\subsection{Results and Discussion}

This section presents the results of applying various clustering algorithms to the terrestrial ecology data. %We first report the results for 10 regions. We then showed that the findings are consistent even when the number of regions is varied.

\begin{figure}[h!]
        \centering
        \begin{subfigure}[b]{0.45\textwidth}
                \includegraphics[width=\textwidth]{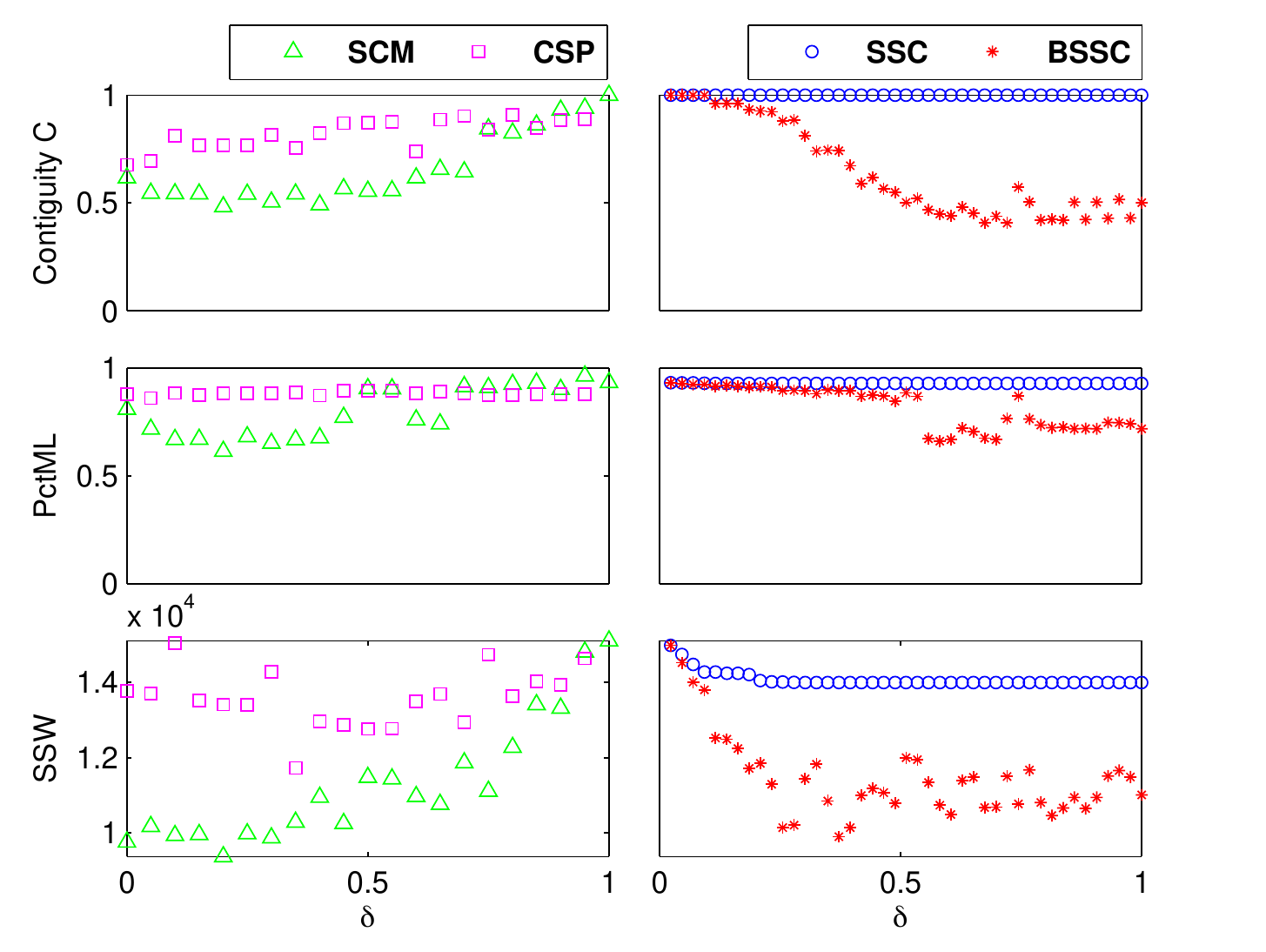}
                \caption{Iowa}
        \end{subfigure}
        \begin{subfigure}[b]{0.45\textwidth}
                \includegraphics[width=\textwidth]{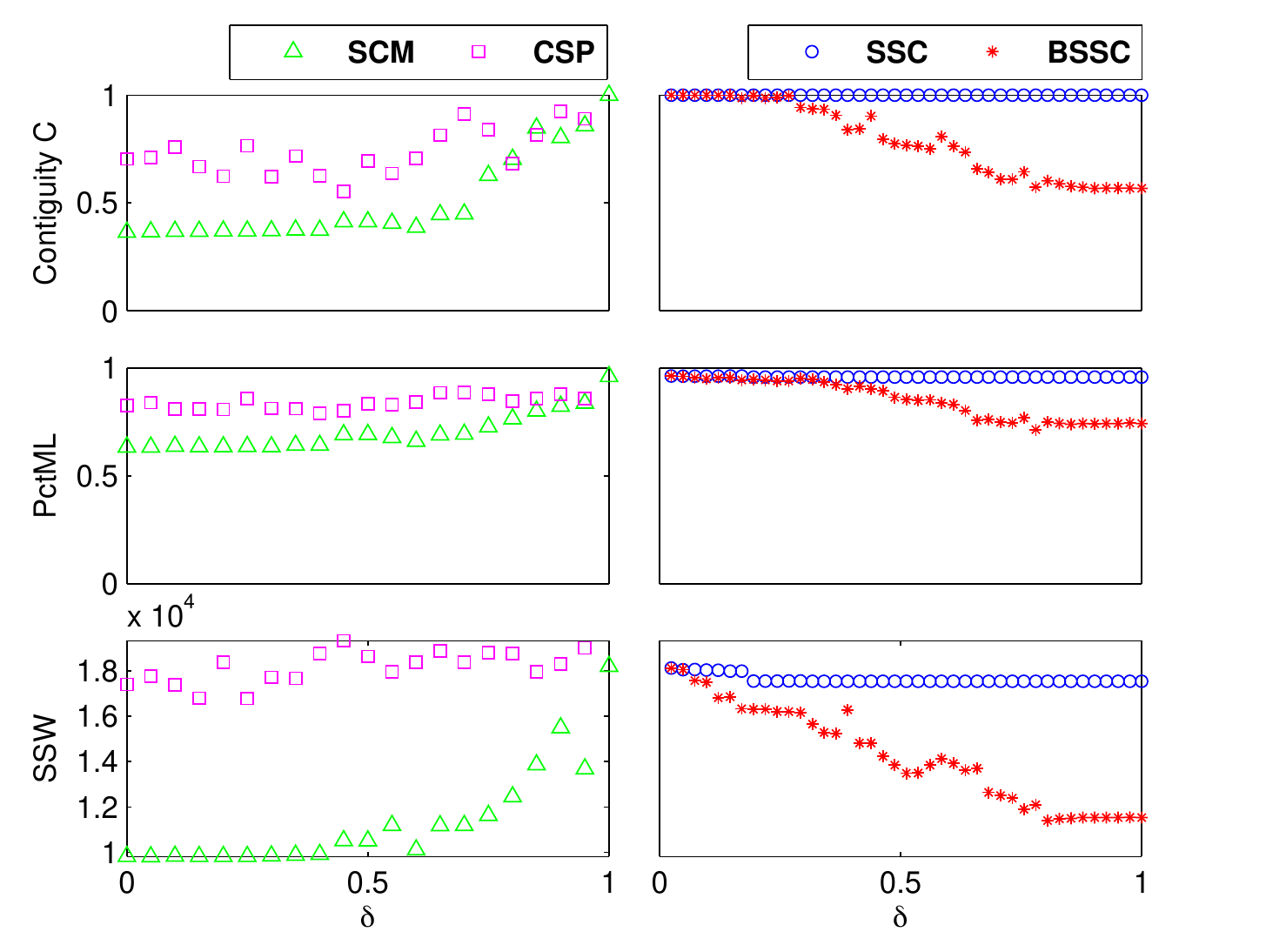}
                \caption{Michigan}
        \end{subfigure}

        \begin{subfigure}[b]{0.45\textwidth}
                \includegraphics[width=\textwidth]{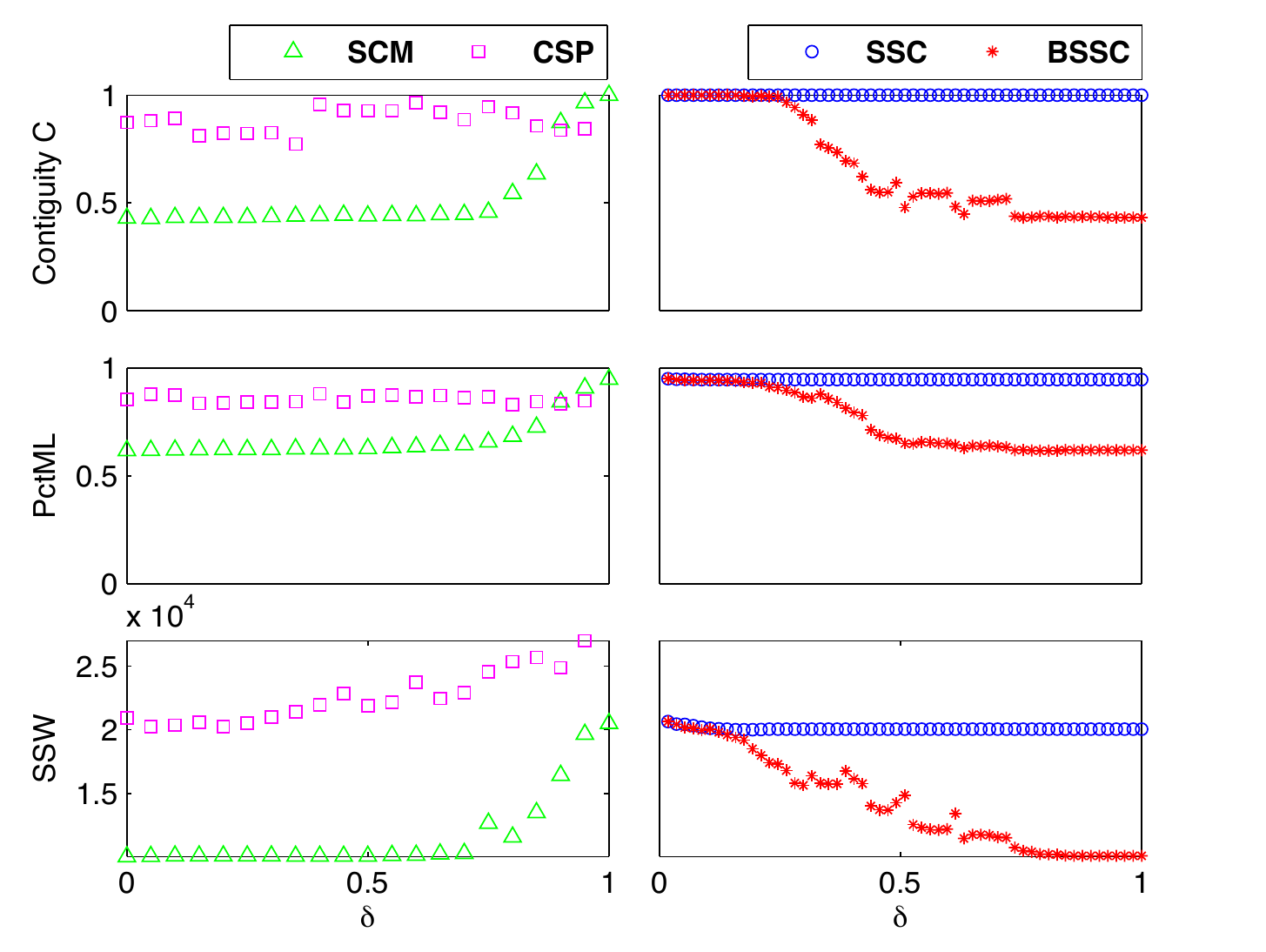}
                \caption{Minnesota}
        \end{subfigure}%
%         \begin{subfigure}[b]{0.46\textwidth}
%                 \includegraphics[width=\textwidth]{NH_parameter}
%                 \caption{New Hampshire}
%         \end{subfigure}
        \caption{Comparison between various constrained spectral clustering algorithms in terms of their landscape homogeneity (SSW) and spatial contiguity (PctML and $c$). The horizontal axis in the plots corresponds to the parameter value $\delta$.}
\label{fig:Parameter_tune}
\end{figure}

\subsubsection{Tradeoff between Homogeneity and Spatial Contiguity}

We first analyze the trade-off between landscape homogeneity and spatial contiguity of the regions by comparing the results for four partitional constrained spectral clustering algorithms: SCM, CSP, SSC, and BSSC. The number of clusters was set to 10. As each algorithm has a parameter $\delta$ that determines whether the clustering should be more biased towards increasing the within-cluster similarity or preserving the ML constraints, we varied the parameter and assessed their performance using the metrics described in Section \ref{sec:metrics}. The $\delta$ parameter for SSC and BSSC has been re-scaled to a range between 0 and 1 by dividing the ML neighborhood size with the diameter of the constraint graph. %The $\beta$ parameter in CSP was also re-scaled as noted in the previous subsection. 

The results are shown in Figure \ref{fig:Parameter_tune}. Observe that the contiguity score ($c$ and PctML) for SCM increases rapidly as $\delta$ becomes closer to 1. This is because increasing $\delta$ would bias the algorithms towards preserving the spatial constraints. A similar increasing trend was also observed for CSP, especially in Iowa and Michigan, though the increase is not as sharp as SCM. In contrast, the contiguity scores would decrease for BSSC as $\delta$ increases because it creates more new ML edges involving spatial units that are not adjacent to each other. For SSC, the contiguity scores do not appear to change by much as $\delta$ increases. This is because the weight $1/k!$ associated with each path of length $k$ decreases rapidly to zero as $k$ increases. As a consequence, the ML neighborhood size for SSC grows until it reaches a maximum size by which increasing $\delta$ will not significantly alter the constraint graph. Thus, SSC is less sensitive to parameter tuning compared to BSSC. Figure \ref{fig:Parameter_tune} also shows there is generally an increasing trend in SSW for SCM and CSP as $\delta$ increases. For SSC, the SSW values do not appear to change significantly with increasing $\delta$ whereas for BSSC, the SSW curve  decreases monotonically as the neighborhood size increases. 

The results of this study showed that the trade-off between landscape homogeneity and spatial contiguity varies among the constrained spectral clustering algorithms. For CSP and SSC, the parameters provided by the algorithms do not allow us to achieve the full range of SSW and contiguity scores. Although these algorithms can produce regions with high contiguity scores, their SSW values were also very high. In contrast, with careful parameter tuning, SCM and BSSC can produce regions with significantly lower SSW compared to CSP and SSC. Observe that the slopes of the curves are steeper near $\delta = 1$ for SCM, which suggests that decreasing $\delta$ below 1 would lead to a dramatic reduction in the contiguity score and SSW of the regions. This makes it harder for SCM to produce regions that are both spatially contiguous and homogeneous. In contrast, the curves for the contiguity scores of BSSC are flatter near $\delta = 0$. This enables the BSSC algorithm to produce regions with homogeneous landscape features yet are still spatially contiguous.

\subsubsection{Performance Comparison for Partitional-based Constrained Clustering}

In this experiment, we set the number of clusters to 10 and selected the $\delta$ parameter that gives the highest contiguity score for each constrained spectral clustering method. If there are more than one parameter values that achieve the highest contiguity score, we chose the one with lowest SSW. For MB, since the Geoclust R package did not support parameter tuning by users, we applied the algorithm using its default setting. 

%---Table--------------------
\begin{table}[t]
\centering
\caption{Performance comparison among various partitional spatially-constrained clustering algorithms with the number of clusters set to 10.}
%Results when number of cluster is equal to 10. Percent of constraint satisfied (\%c), sum of square within cluster(SSW) and relative contiguity metrics (C) are measured for each method on the four study region. SCM-1 is when $\delta = 1$ and SCM-2 is when $\delta = 0.95$.}
\begin{tabular}{|c|l|c|c|c|c|} \hline
%\normalsize
States & Method & PctML & $c$ & SSW& Cbalance \\ \hline \hline
IA &	SCM	&	93.26\%	&	1.00	&	15104 & 0.95	\\
&	CSP	&	87.37\%	&	0.91	&	13628 &0.19	\\
&	MB	&	89.95\%	&	0.69	&	18997 &0.34	\\
&	SSC	&	92.83\%	&	1.00	&	13993&0.95	\\
&	BSSC	&	92.40\%	&	1.00	&	14001&0.94	\\
\hline
MI&	SCM	&	96.08\%	&	1.00	&	18200	&0.85\\
&	CSP	&	87.81\%	&	0.92	&	18307&0.44	\\
&	MB	&	88.76\%	&	0.65	&	16091	&0.91\\
&	SSC	&	95.69\%	&	1.00	&	17534	&0.92\\
&	BSSC	&	94.92\%	&	1.00	&	17485&0.93	\\
\hline
MN &	SCM	&	94.78\%	&	1.00	&	20506	&0.91\\
&	CSP	&	86.62\%	&	0.96	&	23755&0.69	\\
&	MB	&	88.96\%	&	0.64	&	20400&0.67	\\
&	SSC	&	94.57\%	&	1.00	&	19998&0.93	\\
&	BSSC	&	94.12\%	&	1.00	&	19594	&0.91\\
 \hline
% New Hampshire & SCM 	&	84.62\%	&	2533.09	&	1.00\\
% %& SCM ($\delta=0.95$)	&	80.90\%	&	1990.30	&	0.96	\\
% & CSP	&	81.43\%	&	2076.35	&	0.98\\
% & MB	&	88.20\%	&	2978.75	&	0.94\\
% & SSC	&	83.42\%	&	2179.31&	1.00\\
% & BSSC	&	82.89\%	&	1983.86	&	1.00\\\hline
\end{tabular}
\label{table:results}
\end{table}

Table \ref{table:results} summarizes the results of our analysis. SCM, SSC, and BSSC can be tuned to produce regions that are fully contiguous ($c = 1$). The SSW for BSSC and SSC are consistently better than SCM. These results clearly showed the advantage of using a Hadamard product approach instead of a weighted sum approach to integrate spatial constraints into the feature similarity matrix. The limitation of using a weighted sum approach can be explained as follows. Since the highest contiguity score is achieved by setting $\delta = 1$, the clustering solution of SCM is equivalent to applying spectral clustering on the constraint graph only, without considering the feature similarity. If we reduce the parameter value to, say $\delta = 0.95$, its contiguity score decreases sharply (see Figure \ref{fig:Parameter_tune}) while its SSW value is still worse than BSSC. The weighted sum approach has poor SSW because it significantly alters the feature similarity matrix. 

\begin{figure}
 \centering
 \includegraphics[width=45mm]{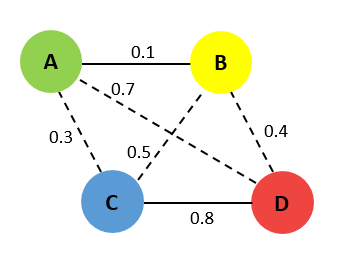}
 \caption{A toy example illustrating the advantage of using Hadamard product for combining constraints with feature similarity. Each labeled node is a data point, with a solid line representing a must-link constraint between two points and a dashed line representing the absence of such constraint. The weight of each edge denotes the feature similarity between two data points.}
 \label{fig:table3}
\end{figure}

To illustrate the limitation of using the weighted sum approach, consider the toy example shown in Figure \ref{fig:table3}. Assume there are 4 data points: A, B, C and D, that need to be clustered. A sample of their pairwise similarity values is shown in Table \ref{tab:example}.
%\vspace{0.3cm}
%\noindent
\begin{table}[h]
\caption{}
\label{tab:example}
\begin{tabular}{|c|c|c|c|c|}\hline
Pairs & Feature & ML  & Weighted & Hadamard \\  
& Similarity & Constraint & Sum & Product \\ \hline
A-B & 0.1 & 1 & 0.955 & 0.1 \\ 
B-C & 0.5 & 0 & 0.025 & 0 \\
C-D & 0.8 & 1 & 0.990 & 0.8\\ \hline
\end{tabular}
\end{table}
%\vspace{0.3cm}

Although the A-B pair has a significantly lower similarity value than C-D, the weighted sum approach inflates the similarity significantly (assuming $\delta=0.95$) which makes it overall similarity to be comparable to C-D. In contrast, the Hadamard product approach simply zeros out the similarity of pairs that do not have ML edges, and thus, will not artificially inflate the similarities of pairs with ML edges. 

%The modified similarity matrix using the weighted sum approach has a significant impact on the resulting regions. 

\begin{figure}
        \centering
 \includegraphics[width=40mm]{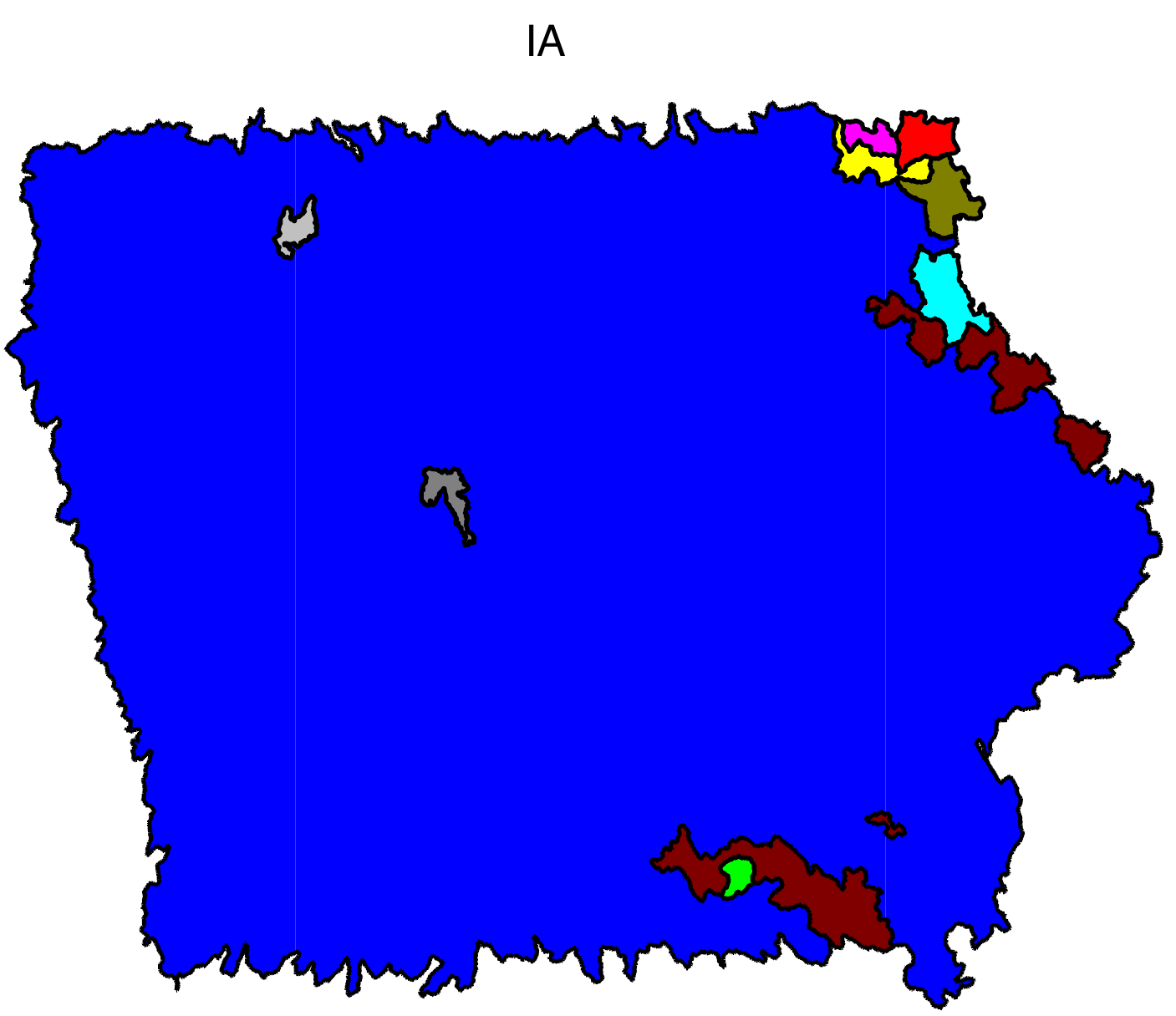}
%                \caption{}
%        \end{subfigure}
         \caption{Regions for Iowa created by the SCM algorithm using the weighted sum approach (with $\delta = 0.95$).}
        \label{fig:lim_sum}
\end{figure}

Furthermore, since the feature similarity is computed using Gaussian radial basis function (see Section \ref{sec:ssc_framework}), the resulting matrix $\mathbf{S}$ for the weighted sum approach is still dense after incorporating the spatial constraints. Unless $\delta=1$, the weighted sum approach will not prevent spatial units that are located far from each other from being placed into the same region. For example, consider the regions found by the weighted sum approach for Iowa, as shown in Figure \ref{fig:lim_sum}. Although the regions appear to be spatially contiguous, they are not compact and have varying sizes. In fact, most of the spatial units were assigned to the same region when $\delta = 0.95$. Even at the lower $\delta$ threshold, its SSW  (14805) is still the worse than the SSW for our framework and CSP. 

\begin{figure}[h]
\centering
    \includegraphics[width = 70mm]{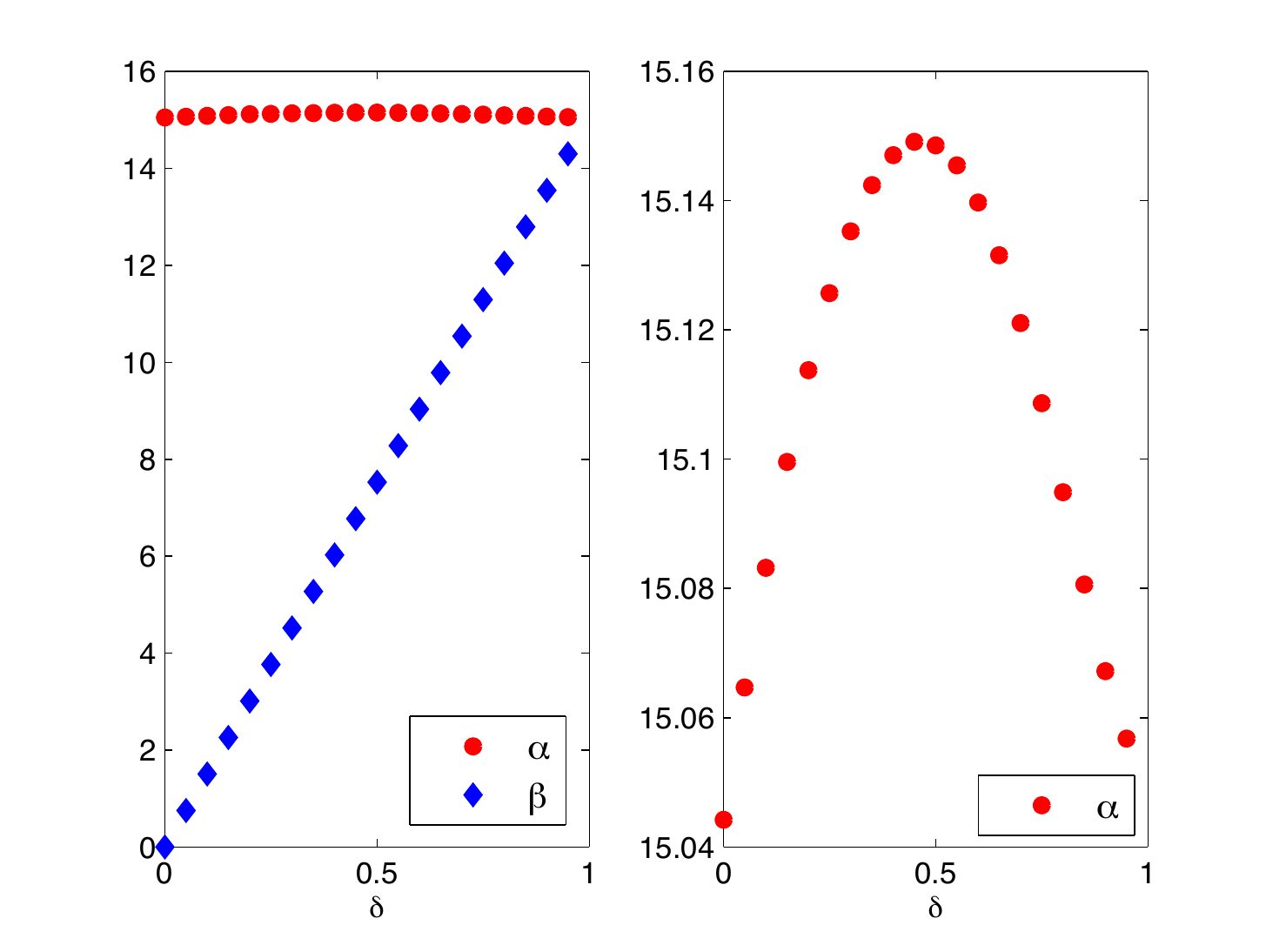}%[trim = 135mm 42mm 5mm 25mm, clip, width=50mm]{toyexample.png}
%    \captionlistentry[table]{A table beside a figure}
%    \captionsetup{labelformat=andtable}
    \caption{The relationship between $\alpha$ and $\beta$ parameter values for the CSP algorithm when applied to a synthetic graph data.}
    \label{fig:alphabeta}
  \end{figure}

The contiguity scores for MB are worse than other constrained clustering methods. Nevertheless, it preserves at least 88\% of the ML edges within the regions. Except for Michigan, its SSW values are also worse than other methods. In contrast, CSP has the lowest contiguity score among all the constrained spectral clustering methods. Except for Iowa, its SSW values are also among the worst. The limitation of CSP~\cite{Wang2010} is a consequence of the parameter used to control its spatial contiguity. As shown in Equation \eqref{eqn:CSP}, the level of spatial constraints satisfied by the clustering solution depends on the parameter $\alpha$. However, instead of directly tuning $\alpha$, the authors suggested to vary another parameter, $\beta$, which was shown to be an upper bound of $\alpha$. The results of our case study showed that increasing the value of $\beta$ does not necessarily imply an increase in $\alpha$. 
%Wang and Davison add a lower bound, $\alpha$, of the satisfaction of the constraint to the objective function. THE LAST PART OF THE PREVIOUS SENTENCE NEEDS REVISING AND THEN THE NEXT SENTENCE IS A FRAGMENT. For constraint matrix C and cluster indicator u. When $C_{ij} = 1$, $u_i$ and $u_j$ are expected to have the same sign, and when $C_{ij} = -1$, $u_i$ and $u_j$ are expected to have different signs. That is to say, when the~\cite{Wang2010} constraint matrix C is well satisfied, $u^TCu = \sum_{i,j} u_iC_{ij}u_j$ will have a larger value. Therefore $u^TCu$ is a metric of how well the constraint is satisfied. Considering the constraint matrix has only must-link, it can be shown that $u^TCu$ has a upper bound $ \frac{vol(C)}{N}$.
%
%From our experiment, we see that controlling $\beta$ cannot control $\alpha$. That is, the increase of $\beta$ in \cite{Wang2010} cannot guarantee the solution is biased towards satisfying the constraint. 
To illustrate this point, we randomly generated a constraint graph that has nine vertices with a randomly generated feature similarity matrix. Assuming the number of clusters is equal to 2, we ran the CSP algorithm with different parameter settings and plotted their values of $\alpha$ and $\beta$ in Figure \ref{fig:alphabeta}. Although this figure shows that the value of $\beta$ (blue diamond) is a lower bound of $\alpha$ (red circle), the bound is so loose that it can not guarantee that increasing $\beta$ will increase $\alpha$. In fact, the figure on the right  shows that $\alpha$ is not a monotonically increasing function of $\beta$. This is why controlling its parameter value will not always guarantee that the regions will be contiguous even when $\delta = 1$ (unlike SCM and the Hadamard product approaches).

%We notice that, when SCM $\delta = 1$, it is equivalent to clustering on constraint only and no feature similarity will be considered; thus, we report the results of when $\delta = 1$ and the result when $\delta$ close to 1 (in our case $\delta = 0.95$) in Table\ref{table:results}. We can see from the table that the contiguity metric for SCM is worse than the other methods when $\delta \neq 1$. Among these 4 methods, SSC always gives contiguous clusters while maintaining a relatively low SSW. 

%-------------region results visualization-------------------------------------------------
\begin{figure*}[h!]
        \centering
        \begin{subfigure}[b]{0.18\textwidth}
                \includegraphics[trim = 30mm 0mm 0mm 20mm,clip, width=\textwidth]{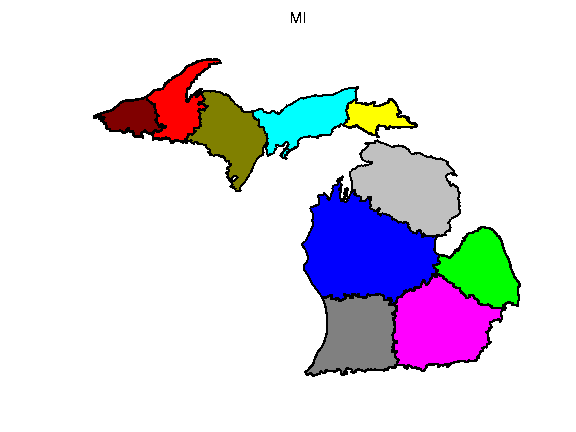}
                \caption{SCM}
                \label{fig:scmMI}
        \end{subfigure}
        \begin{subfigure}[b]{0.18\textwidth}
                \includegraphics[trim = 30mm 0mm 0mm 20mm,clip, width=\textwidth]{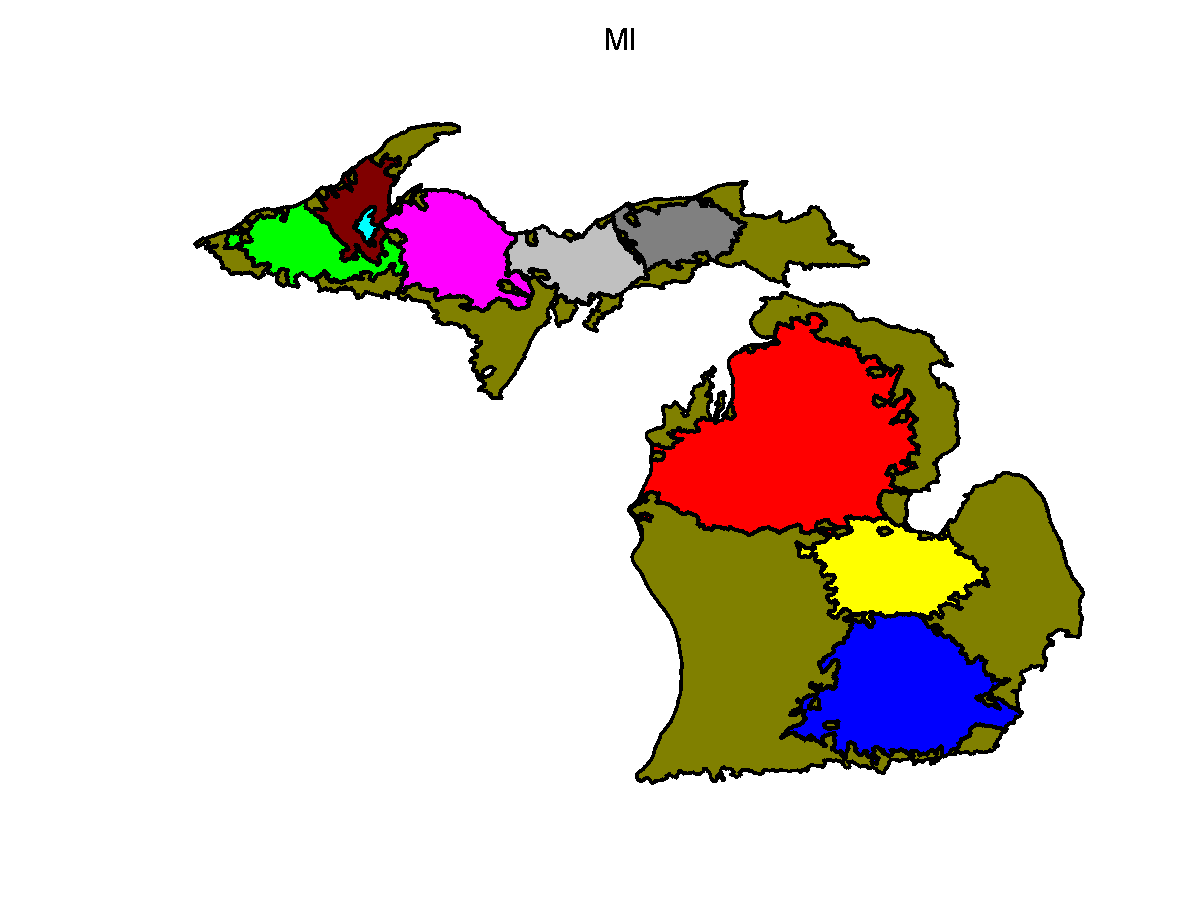}
                \caption{CSP}
                \label{fig:cspMI}
        \end{subfigure}              
        \begin{subfigure}[b]{0.18\textwidth}
                \includegraphics[trim = 22mm 0mm 0mm 20mm,clip, width=\textwidth]{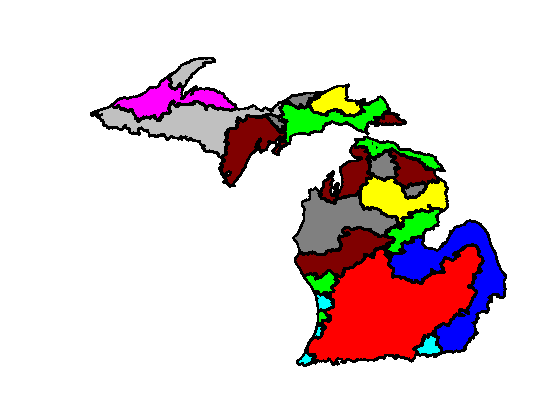}
                \caption{MB}
                \label{fig:geoMI}
        \end{subfigure}       
        \begin{subfigure}[b]{0.18\textwidth}
                \includegraphics[trim = 30mm 0mm 0mm 20mm,clip, width=\textwidth]{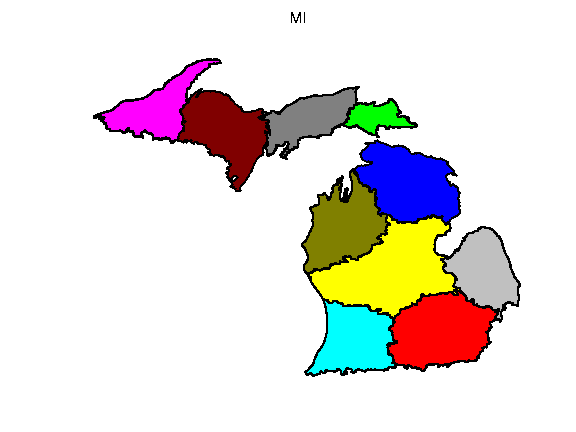}
                \caption{SSC}
                \label{fig:ccspMI}
        \end{subfigure}       
        \begin{subfigure}[b]{0.18\textwidth}
                \includegraphics[trim = 30mm 0mm 0mm 20mm,clip, width=\textwidth]{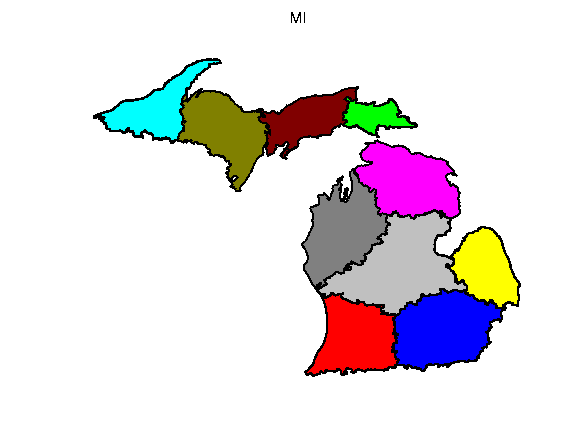}
                \caption{BSSC}
                \label{fig:ccsp1MI}
        \end{subfigure}
        
        \caption{Regions created by the SCM, CSP, MB, SSC and BSSC algorithms for the state of Michigan.}
        \label{fig:BSSC}
\end{figure*}

In terms of the Cbalance measure, our results suggest that SCM, SSC, and BSSC achieve the highest cluster balance for all three states. This can be verified by examining the regions generated by all the competing algorithms for the the state of Michigan, as shown in Figure \ref{fig:BSSC}\footnote{The corresponding maps of regions for other states can be found in \cite{Yuan2015}.} As can be seen from the figure, the regions produced by SCM, SSC, and BSSC are more compact and uniform in size compared to CSP and MB. However, the SSW for SCM is worse than the SSW for our proposed SSC and BSSC algorithms. This is not surprising as SCM cannot produce contiguous clusters unless $\delta$ is very close to 1. If $\delta$ is lowered slightly to 0.95, the regions changed significantly, as shown in Figure \ref{fig:lim_sum}.
%This is consistent with the fact that all three algorithms produce regions that are most contiguous. 
%for each method. SCM can produce cluster with high cluster balance. But as we demonstrated before, SCM cannot produce contiguous cluster unless $\delta$ is 1 or almost 1. In that case, 
By setting $\delta$ close to 1, SCM will focus only on preserving the spatial constraints, and thus, has worse cluster homogeneity compared to our algorithms. %For CSP and MB, the cluster balance is low. And our proposed SSC and BSSC achieves good cluster balance in all three states}
%Finally, the regionalization system generated by all the competing algorithms for Michigan are shown in Figure \ref{fig:BSSC}. \SY{For visualization for all 3 study regions see \cite{Yuan2015}.}  For SCM, although the regions are contiguous, their SSW values are worse than SSC and BSSC. 
%For Iowa and Michigan, the region boundaries for SSC and BSSC are almost identical. 
Thus, our results clearly show the benefits of applying BSSC to develop homogeneous and spatially contiguous regions compared to other baseline algorithms. These results hold true even when the number of regions is varied. A comparison of the results for different
number of clusters can be found in our earlier work \cite{Yuan2015}.

%\subsubsection{Limitations on SCM and CSP}

%The SCM method solved the spectral clustering objective function with a weighted sum of feature similarity and constraint. One drawback of this method is that the feature similarity from RBF will be a dense matrix. Thus, W based on linear combination of S and C will be a dense graph. I AM NOT SURE IF YOU HAVE USED THESE ABBREVIATIONS ALREADY IN THE MS - SEEM NEW TO ME IN THE PAST SENTENCE. This result will not forbid data points far away from being in the same cluster. That is, unless $\delta =1$, the cluster cannot be guaranteed to be contiguous. 

\subsubsection{Performance Comparison for Hierarchical-based Constrained Clustering}

In this section, we compared our proposed HSSC algorithm against the spatially-constrained agglomerative clustering methods for constructing nested regions. %Figure \ref{fig:HMI_example} shows an example of the nested regions obtained for Michigan as we vary the number of regions from 2 to 10. When the number of regions is equal to 2, the algorithm separates the region into two disjoint areas. When $K = 3$, one of regions is selected to be further split into two smaller subregions. 
%We compare the three types of measurements for HSSC and other four baseline algorithms described in Section \ref{subsec:baseline} for 3 study region. 
Note that all of the algorithms apply a Hadamard product to combine the constrained matrix $\mathbf{S}_c$ (for a given $\delta$) with the feature similarity matrix $\mathbf{S}$ to generate the combined matrix $\mathbf{S}^{\textrm{total}}$ before applying hierarchical clustering. For the spatially-constrained agglomerative hierarchical clustering methods, the regions are iteratively merged starting from the initial $\mathbf{S}^{\textrm{total}}$. 

\begin{figure*}
        \centering
        \begin{subfigure}[b]{1.1\textwidth}
                \includegraphics[trim = 30mm 100mm 10mm 50mm,clip, width=\textwidth]{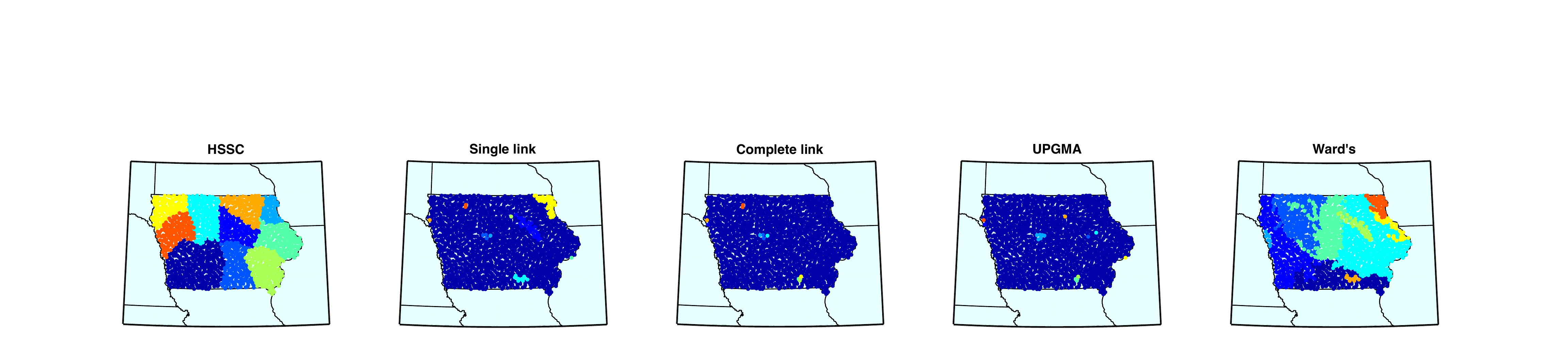}
                \caption{Regions in IA developed by 5 hierarchical algorithm with 10 clusters}
                \label{fig:HIA}
        \end{subfigure}
        \begin{subfigure}[b]{1.1\textwidth}
                \includegraphics[trim = 30mm 90mm 10mm 60mm,clip, width=\textwidth]{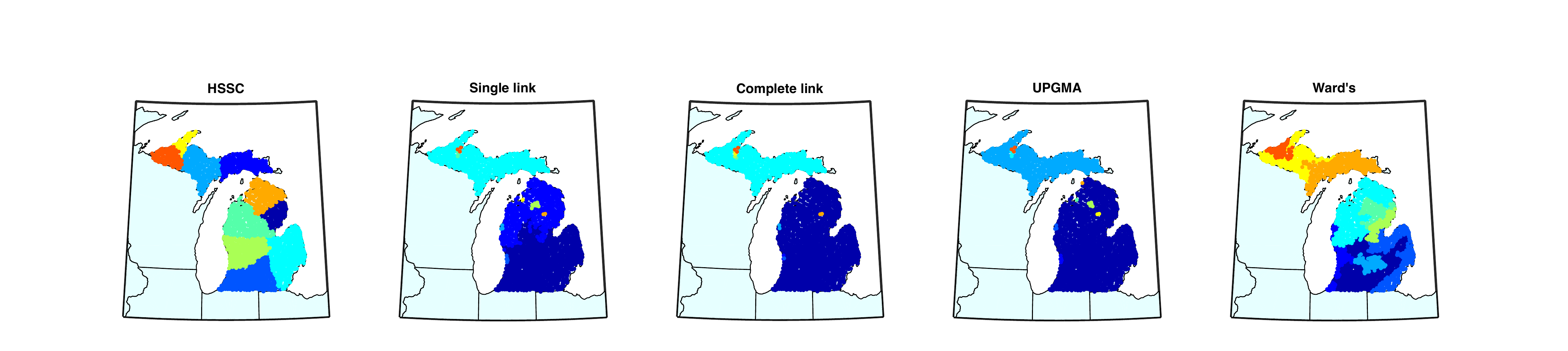}
                \caption{Regions in MI developed by 5 hierarchical algorithm with 10 clusters}
                \label{fig:HMI}
        \end{subfigure}              
        \begin{subfigure}[b]{1.1\textwidth}
                \includegraphics[trim = 30mm 90mm 10mm 60mm,clip, width=\textwidth]{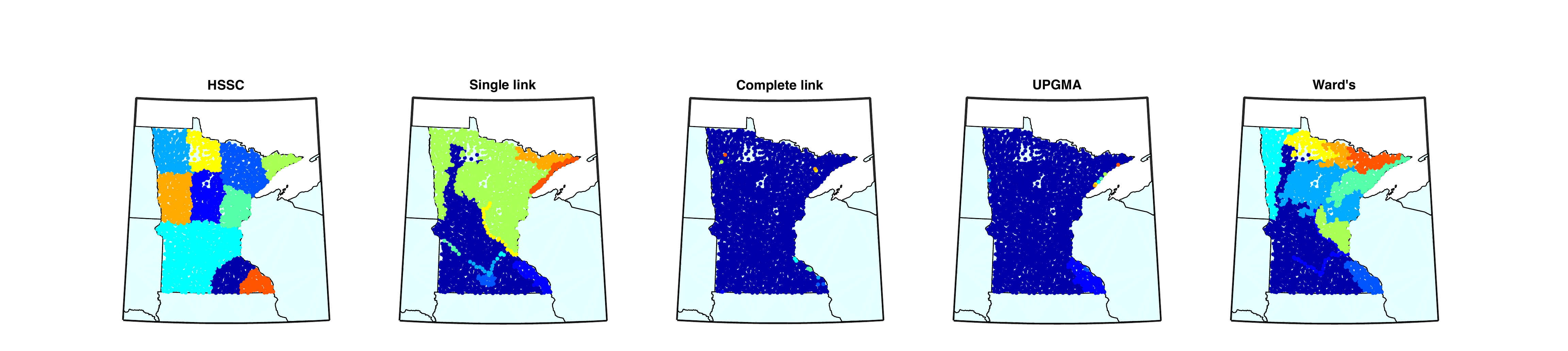}
                \caption{Regions in MN developed by 5 hierarchical algorithm with 10 clusters}
                \label{fig:HMN}
        \end{subfigure} 

        \caption{Regions developed by 5 hierarchical spatially constrained clustering algorithm for 3 study regions}
        \label{fig:Hie}
\end{figure*}

For a fair comparison, we set $\delta = 1$ for all the methods. The results for $k=10$ are summarized in Table \ref{table:Hresults}. In terms of region contiguity, observe that all the methods can achieve $c=1$.  However, the spatially constrained complete link and UPGMA algorithms produce the highest PctML values while the Ward's method produces the lowest value. The PctML for our proposed HSSC algorithm is still relatively high and comparable to its non-hierarchical counterparts, SSC and BSSC (see Table \ref{table:results}). Despite their high spatial contiguity, both the spatially-constrained complete link and UPGMA methods have the worst SSW compared to other methods. Worst still, their Cbalance values are close to 0, suggesting that the sizes of their regions are highly imbalanced.  This can be seen from the maps shown in Figure \ref{fig:Hie}, where there is a large region covering the majority of the landscape in each state. In contrast, our HSSC algorithm has the highest Cbalance, consistently producing regions that are compact and approximately similar in sizes.  

The spatially-constrained single link method has comparable PctML but slightly lower SSW compared to HSSC. It also suffers from the same imbalance region problem as the complete link and UPGMA methods. 
%As we can see from the table all methods can produce contiguous clusters(c = 1) and there's a trade off between the PctML and SSW. Among all the methods, our proposed HSSC achieves best cluster balance for all three states, This is because HSSC is a top-down algorithm that start from one cluster and keep splitting until you get N clusters. At each step, HSSC choose the cluster with largest SSW to split into two. Large clusters tend to hae larger SSW and thus they often get to be split. In addition, spectral clustering objective function is designed to produce balanced clusters. Hierarchical clustering with 
Meanwhile, the spatially-constrained Ward's method achieves the lowest SSW among all the competing methods, which is not surprising since the algorithm is designed to minimize the SSW in each iteration of the algorithm. However, this comes at the expense of its poor PctML values, which is the worst among all the competing methods. In addition, the Ward's method is known to suffer from the cluster inversion problem \cite{murtagh1985}, in which its objective function is not monotonically non-decreasing as the number of clusters increases. In short, our HSSC algorithm outperforms the complete link, UPGMA, and Ward's methods in 2 of the 3 evaluation criteria. Its PctML and SSW is also quite similar to single link, which suffers from the region imbalanced problem. 

%Hierarchical clustering with Complete link achieves best PctML for all three states. 
%For each state, if we compare pairwise method,
%our proposed HSSC method is better than each of the existing hierarchical methods in 2 out of 3 criteria. 
%HSSC is outperforms completelink and UPGMA in cluster balance and SSW. HSSC is better than Ward's method in cluster balance and PctML.

%---Table--------------------
\begin{table}[t]
\centering
\caption{Performance comparison among various hierarchical spatially-constrained clustering algorithms with $\delta = 1$ and the number of clusters set to 10.}
\begin{tabular}{|c|l|c|c|c|c|} \hline
States & Method & PctML & $c$ & SSW& Cbalance \\ \hline \hline
IA	&	HSSC	&	92.53	\%	&	1	&	15080	&	0.92	\\
	&	Single link	&	96.02	\%	&	1	&	14191	&	0.04	\\
	&	Complete link	&	98.75	\%	&	1	&	18309	&	0.02	\\
	&	UPGMA	&	98.45	\%	&	1	&	18227	&	0.02	\\
	&	Ward's	&	84.96	\%	&	1	&	9281	&	0.48	\\
    \hline
MI	&	HSSC	&	95.41	\%	&	1	&	17420	&	0.86	\\
	&	Single link	&	95.31	\%	&	1	&	16575	&	0.08	\\
	&	Complete link	&	98.72	\%	&	1	&	20083	&	0.03	\\
	&	UPGMA	&	98.33	\%	&	1	&	19441	&	0.04	\\
	&	Ward's	&	86.28	\%	&	1	&	14047	&	0.79	\\
\hline
MN	&	HSSC	&	95.02	\%	&	1	&	20075	&	0.82	\\

	&	Single link	&	90.70	\%	&	1	&	19660	&	0.20	\\
	&	Complete link	&	99.17	\%	&	1	&	33431	&	0.01	\\
	&	UPGMA	&	98.81	\%	&	1	&	28681	&	0.02	\\
	&	Ward's	&	87.69	\%	&	1	&	14183	&	0.63	\\

\hline
\end{tabular}
\label{table:Hresults}
\end{table}

%We show the clustering results of all five methods for 3 study regions with 10 clusters on map in Figure \ref{fig:HIA}, \ref{fig:HMI} and \ref{fig:HMN}. As we can see, our HSSC tends to produce balance-sized and regular-shaped clusters. Ward's method can produce relatively balanced-sized irregular-shaped clusters. And the rest methods performs worst in terms of the cluster balance.
%---------------------------------------
% -----------------------------------------------
\begin{figure*}
        \centering 
        \begin{subfigure}[b]{1.1\textwidth}
        \includegraphics[trim = 10mm 30mm 10mm 30mm,clip, width=\textwidth]{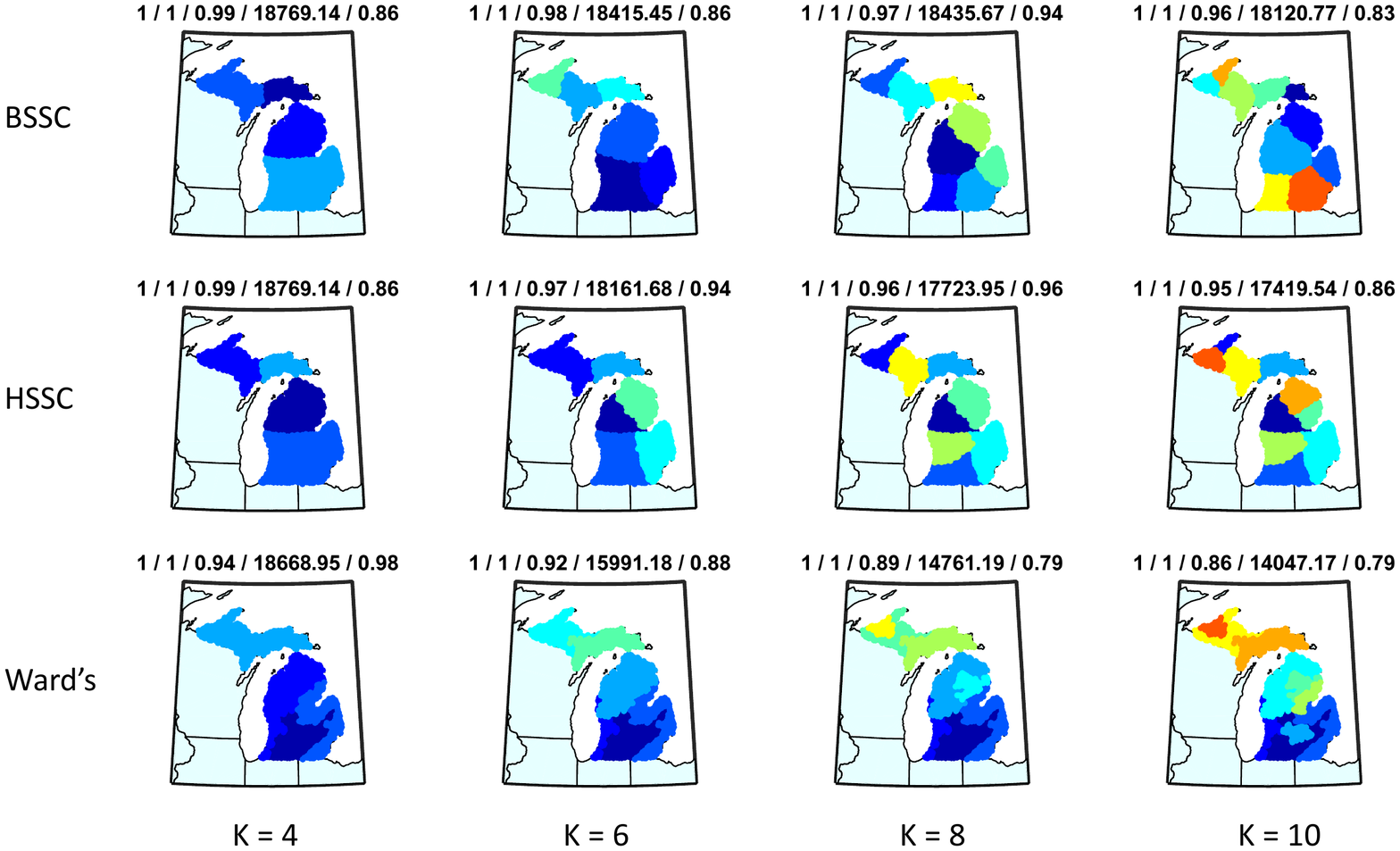}    
        \caption{Results for $\delta = 1/41$}
        \label{fig:HMIdelta1}
		\end{subfigure}
        
        \begin{subfigure}[b]{1.1\textwidth}
        \includegraphics[trim = 10mm 30mm 10mm 30mm,clip, width=\textwidth]{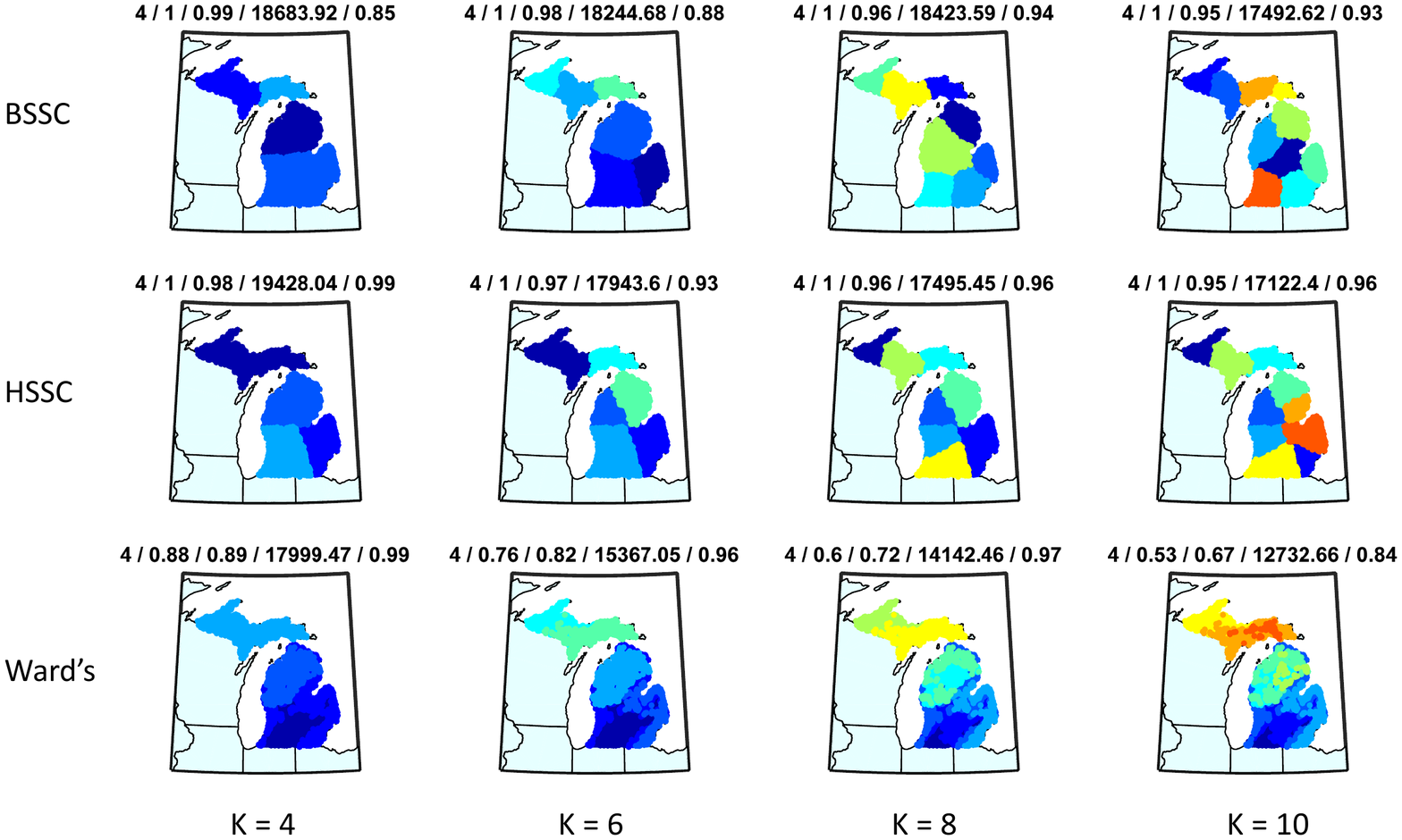}    
        \caption{Results for $\delta = 4/41$}
        \label{fig:HMIdelta4}
        \end{subfigure}
        \caption{Comparison between BSSC, HSSC and Ward's method for number of cluster $K = 4, 6, 8, 10$. The five metrics evaluated are listed on top of each figure, namely: unnormalized delta, contiguity metric c, PctML preserved, SSW and cluster balance. Results for $\delta = 1/41$ are shown in (a) and results of the unnormalized $\delta = 4/41$ are shown in (b). }
\end{figure*}

Figures \ref{fig:HMIdelta1} and \ref{fig:HMIdelta4} show a comparison between the regions produced by BSSC, HSSC, and the Ward's method for the state of Michigan as we vary the number of regions from 4 to 10. We show the value of the unnormalized $\delta$ along with four metrics---$c$, PctML, SSW, and Cbalance---at the top of each diagram. Recall that the normalized $\delta$ is the ratio between the original $\delta$ given in Equation \eqref{eqn:constraints} and the diameter of the spatial constraint graph. As we increase $\delta$ from 1/41 to 4/41, the Ward's method no longer produces regions that are contiguous unlike the BSSC and HSSC methods. The Cbalance for Ward's method is also worse than our algorithms except when the number of clusters is small. The results for BSSC is quite comparable to HSSC, since both of them are based on the same spatially constrained spectral clustering framework. The Cbalance and SSW for HSSC are slightly better than BSSC but its PctML is slightly worse. While this may seem counter-intuitive since BSSC directly optimizes the objective function for spatially-constrained spectral clustering whereas HSSC uses a greedy recursive bisection strategy, it is worth noting that the objective function does not depend solely on the feature similarities within the regions. Instead, it takes into account the spatial constraint matrix $C$ as well.   
In fact, if we compare the values of the objective functions for BSSC and HSSC at $k=10$ for the state of Michigan, BSSC has a noticeably lower value (13458.13) compared to HSSC (14284.21). %This result is consistent with the theorem in \ref{th:th1}

%We then compare the best non-hierarchical method BSSC with HSSC and Wards method for different number of clusters for $\delta = 1$ in Figure \ref{fig:HMIdelta1} and for $\delta = 4$ in Figure \ref{fig:HMIdelta4}. The five metrics are top of each figure are: delta, contiguity metric c, PctML preserved, SSW and cluster balance. As we increase $\delta$ from 1 to 4, Ward's method failed to produce contiguous clusters. As we increase the number of clusters, the results of HSSC is close the results of BSSC. At last, we compare how sensitive the three methods are with respect to the parameter $\delta$. 

Finally, we also compare the stability of the regions as we increase ML neighborhood size ($\delta$). For this experiment, we show the results for the state of Michigan (in which the diameter of the constraint graph is 41) and varies the normalized $\delta$ from 1/41 to 4/41. We use the adjusted rand index~\cite{rand1971} to compare the similarity between two clustering results. A high adjusted rand index would suggest that the regions found are stable, i.e., do not vary significantly with different values of $\delta$.  Table \ref{table:adj} shows the mean adjusted rand index (averaged over the number of clusters, which varies from 1 to 10) for HSSC, BSSC, and Ward's method. The results suggest that  the proposed BSSC and HSSC methods are less sensitive to the change in $\delta$ compared to the Ward's method, which is another advantage of using our frameworks.

\begin{table}[h]
\centering
\begin{tabular}{|c|c|c|c|} \hline
 &BSSC & HSSC & Ward's   \\ \hline 
Mean Adjusted Rand Index	&0.92	&0.86	&0.66\\
\hline
\end{tabular}
\caption{Stability of the regions generated by different hierarchical clustering methods for the state of Michigan. The mean Adjusted Rand Index is computed for each method by comparing the similarity between the regions found with $\delta = 1/41$ to the regions found with $\delta = 4/41$.}
\label{table:adj}
%\vspace{-1cm}
\end{table}

%The last point we want to mention is that when you compare the HSSC results with BSSC results. We notice that the SSW of BSSC is better in IA and MN but worse in MI than the results in HSSC. Although BSSC, a non-hierarchical model, optimize the objective function(Equation \ref{eqn:spectralcluster}) directly, while hierarchical model HSSC bisecting at every step, the overall objective function is not solely optimizing SSW. Therefore, it is likely SSW of HSSC outperforms BSSC. But if we compare the objective function value for the two methods in MI, BSSC is better than HSSC(13458.13 for BSSC and 14284.21 for HSSC). 

%--------------------------------------------------
\section{Conclusions} \label{sec:con}
This research investigated the feasibility of applying constrained spectral clustering to the regionalization task. We compared several constrained spectral clustering methods and showed the trade-off between landscape homogeneity and spatial contiguity of their resulting regions.  We presented two algorithms, SSC and BSSC, that uses a Hadamard product approach to combine the similarity matrix of landscape features 
%{\color{red} [PAS: IS A MORE COMMON TERM 'CLUSTER HOMOGENEITY? IF SO, ALWAYS USE THAT THROUGHOUT THE MS AND IN THE TITLE??]} 
with spatial contiguity constraints. The results of our case study showed that the proposed BSSC algorithm is most effective in terms of producing spatially contiguous regions that are homogeneous. The extension of this algorithm to a hierarchical clustering setting also shows its advantages in producing regions that are more balanced in size compared to other hierarchical spatially-constrained algorithms. It also achieves high spatial contiguity and moderate SSW, comparable to the results of its non-hierarchical counterpart (BSSC). 

\section{Acknowledgements} \label{sec:acknowledgments}
This research was funded in part by the National Science Foundation under grant \#EF-1065786 and \#IIS-1615612.

\section{Conflict of Interest}
On behalf of all authors, the corresponding author states that there is no conflict of interest. 
\bibliographystyle{spmpsci}      % mathematics and physical sciences

%\begin{thebibliography}{}
% Author, Article title, Journal, Volume, page numbers (year)
% Author, Book title, page numbers. Publisher, place (year)
% \end{thebibliography}

\end{document}